\pdfoutput=1
\documentclass{article}

\usepackage{iclr2027_conference,times}

\usepackage{amsmath,amsfonts,bm}

\def\eqref#1{equation~\ref{#1}}

\def\1{\bm{1}}

\DeclareMathAlphabet{\mathsfit}{\encodingdefault}{\sfdefault}{m}{sl}
\SetMathAlphabet{\mathsfit}{bold}{\encodingdefault}{\sfdefault}{bx}{n}

\usepackage{graphicx}
\usepackage{algorithm}
\usepackage{algpseudocode}
\usepackage{hyperref}
\usepackage{url}
\usepackage{amssymb}
\usepackage{booktabs}
\usepackage{xcolor}
\usepackage{caption}
\usepackage[T1,OT1]{fontenc}
\usepackage{listings}
\usepackage{textcomp}

\definecolor{darkblue}{rgb}{0, 0, 0.5}
\hypersetup{colorlinks=true, citecolor=darkblue, linkcolor=darkblue, urlcolor=darkblue}

\title{Match the Canvas, Not Just the Answer:\\
Predictability-Aware SFT for\\
Diffusion Language Models}

\author{%
\textbf{Miao Li}\textsuperscript{1,*}
\quad
\textbf{Hanyang Jiang}\textsuperscript{1,*}
\quad
\textbf{Hengyu Fu}\textsuperscript{2}
\quad
\textbf{Baihe Huang}\textsuperscript{2}
\\
\textbf{Yating Liu}\textsuperscript{3}
\quad
\textbf{Yuhang Cai}\textsuperscript{2}
\quad
\textbf{Changkun Guan}\textsuperscript{1}
\quad
\textbf{Zhongzhu Chen}\textsuperscript{4}
\\
\textbf{Jiecheng Lu}\textsuperscript{1}
\quad
\textbf{Michael A. Boateng}\textsuperscript{1}
\quad
\textbf{Pascal Van Hentenryck}\textsuperscript{1}
\\[0.5em]
\textsuperscript{1}Georgia Institute of Technology
\quad
\textsuperscript{2}University of California, Berkeley
\\
\textsuperscript{3}University of Chicago
\quad
\textsuperscript{4}University of Michigan
\\[0.3em]
\texttt{\{mli746,scottjhy,cguan62,jlu414\}@gatech.edu}
\\
\texttt{\{mboateng6,pvh\}@gatech.edu}
\quad
\texttt{\{hengyuf,baihe\_huang\}@berkeley.edu}
\\
\texttt{willcai@berkeley.edu}
\quad
\texttt{yatingliu@uchicago.edu}
\quad
\texttt{zhongzhc@umich.edu}
\\[0.3em]
\textsuperscript{*}Equal contribution.
}

\iclrfinalcopy

\begin{document}

\maketitle
\fancyhead{}
\renewcommand{\headrulewidth}{0pt}

\begin{abstract}
Diffusion language models (DLMs) enable flexible parallel generation by resolving tokens according to the model's evolving predictability rather than a fixed left-to-right order. This flexibility, however, introduces training choices absent from standard autoregressive teacher forcing: which reference tokens are visible when a target is supervised and how much generation space is allocated. Standard supervised fine-tuning (SFT) typically makes these choices through random masking, independently of inference. Our analysis shows how this training--inference mismatch can misdirect supervision. Random masking
can conceal prediction errors or penalize uncertainty that decoding would otherwise resolve, while the attention-visible input length can change which response formulations the model favors, even when the underlying
solution is unchanged.
Motivated by these findings, we propose \emph{Predictability-Aware Canvas Selection and Supervision (PACS)}, an SFT method. PACS uses the model's own predictability to select each example's attention-visible input length and supervises reference tokens immediately before their confidence-guided revelation. Across reasoning and code-generation benchmarks, PACS consistently achieves the highest mean accuracy among the compared SFT methods, with gains of up to 12.84 percentage points over vanilla SFT on individual benchmarks. 

\end{abstract}
\noindent\textbf{GitHub:} \href{https://anonymous.4open.science/r/pacs-FB4E/README.md}{\nolinkurl{anonymous.4open.science/r/pacs-FB4E/README.md}}\par
\section{Introduction}
\label{sec:introduction}
Diffusion language models (DLMs) offer a promising route to parallel text generation
\citep{lou2024discrete,sahoo2024simple,shi2024simplified}.
Their flexible generation order allows tokens to be generated in parallel according to the model's evolving predictability rather than a fixed left-to-right schedule.

\begin{figure}
    \centering    \includegraphics[width=0.9\linewidth]{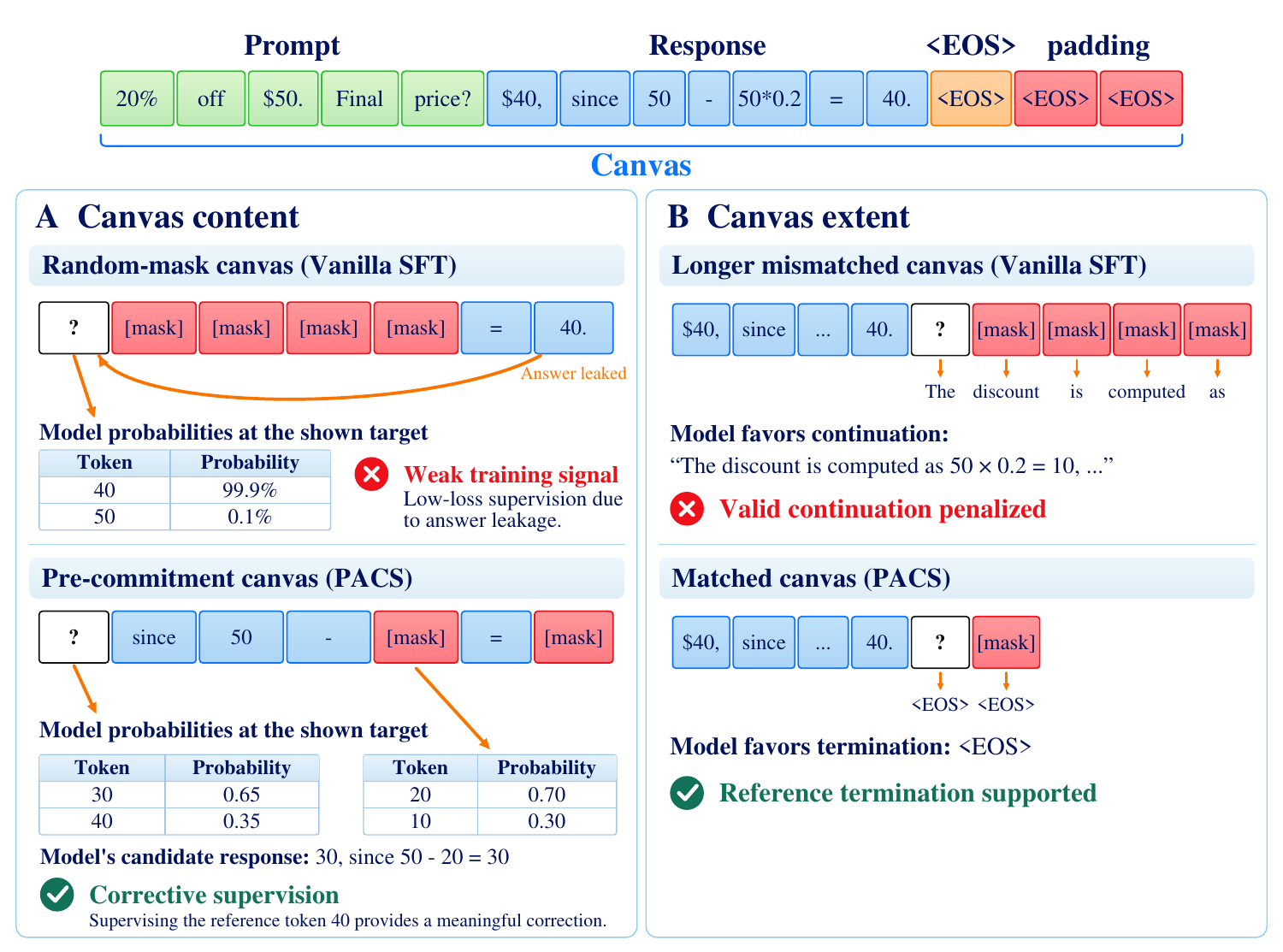}
    \caption{
    \textbf{The training canvas changes reference supervision.}
    \textbf{Top:} A canvas includes prompt tokens (green), visible answer
    tokens (blue), and masked positions (red).
    \textbf{A:} Visible reference content can conceal or expose a
    commitment-time disagreement.
    \textbf{B:}
    Canvas extent can alter the prediction problem, inducing an extra stopping signal on a mismatched canvas but not on the
    matched canvas.}
    \label{fig:intro}
\end{figure}

At inference time, a widely used way to realize this flexibility is \emph{confidence-guided decoding}, which uses token-level confidence to decide which predictions to commit \citep{wu2025fastdllm}.
Standard autoregressive (AR) decoding extends a contiguous prefix one token at a time. Masked DLMs such as LLaDA \citep{nie2025large} and Dream \citep{ye2025dream}, by contrast, begin with an allocated response region filled with $\mathtt{[MASK]}$ tokens.
We call the full attention-visible input---the prompt, visible response tokens, and remaining masked positions---the \emph{canvas}, illustrated in Fig.~\ref{fig:intro}. As confident predictions replace selected masks, the canvas evolves and
conditions subsequent predictions.

At training time, this flexibility makes DLM SFT inherently more difficult to specify. In AR teacher forcing, once a reference answer is given, the context for predicting each target is fixed by the preceding reference tokens. For a masked DLM, however, the reference fixes the target tokens but does not itself specify the canvas from which they should be predicted.
The same reference can therefore be paired with different canvas lengths, visible reference tokens, and mask patterns, each defining a different prediction problem.

Full-attention masked DLMs such as LLaDA and Dream accommodate variable-length responses within a preallocated generation canvas, with $\mathtt{EOS}$ marking the response boundary.
During SFT, their training recipes pad shorter examples with the
same $\mathtt{EOS}$ token for batched computation.
These positions remain attention-visible and participate in masking and reconstruction, teaching the model to predict $\mathtt{EOS}$ beyond the response.
The allocated extent therefore contributes to the context in which
the reference is predicted.
As we show, this choice can introduce training signals that adjust
how a correct solution is expressed, rather than correct its reasoning.

Vanilla masked-diffusion SFT typically constructs the visible context by randomly masking the reference response, independently of the model's confidence-guided decoding decisions.
During inference, by contrast, those decisions determine which
predictions are committed and hence what context becomes available next.
We therefore study how the model's own decoding behavior can guide
training-canvas construction:
\begin{center}
\emph{A reference answer tells us which tokens to teach---but how
can the model's own predictions help us choose the training canvas?}
\end{center}
We investigate how the canvas's revealed content and length affect
supervision for the same reference.
These choices determine which reference tokens are visible when a
target is supervised and how many input positions the model can
attend to.

\begin{figure}[t]
    \centering
    \includegraphics[width=0.6\linewidth]{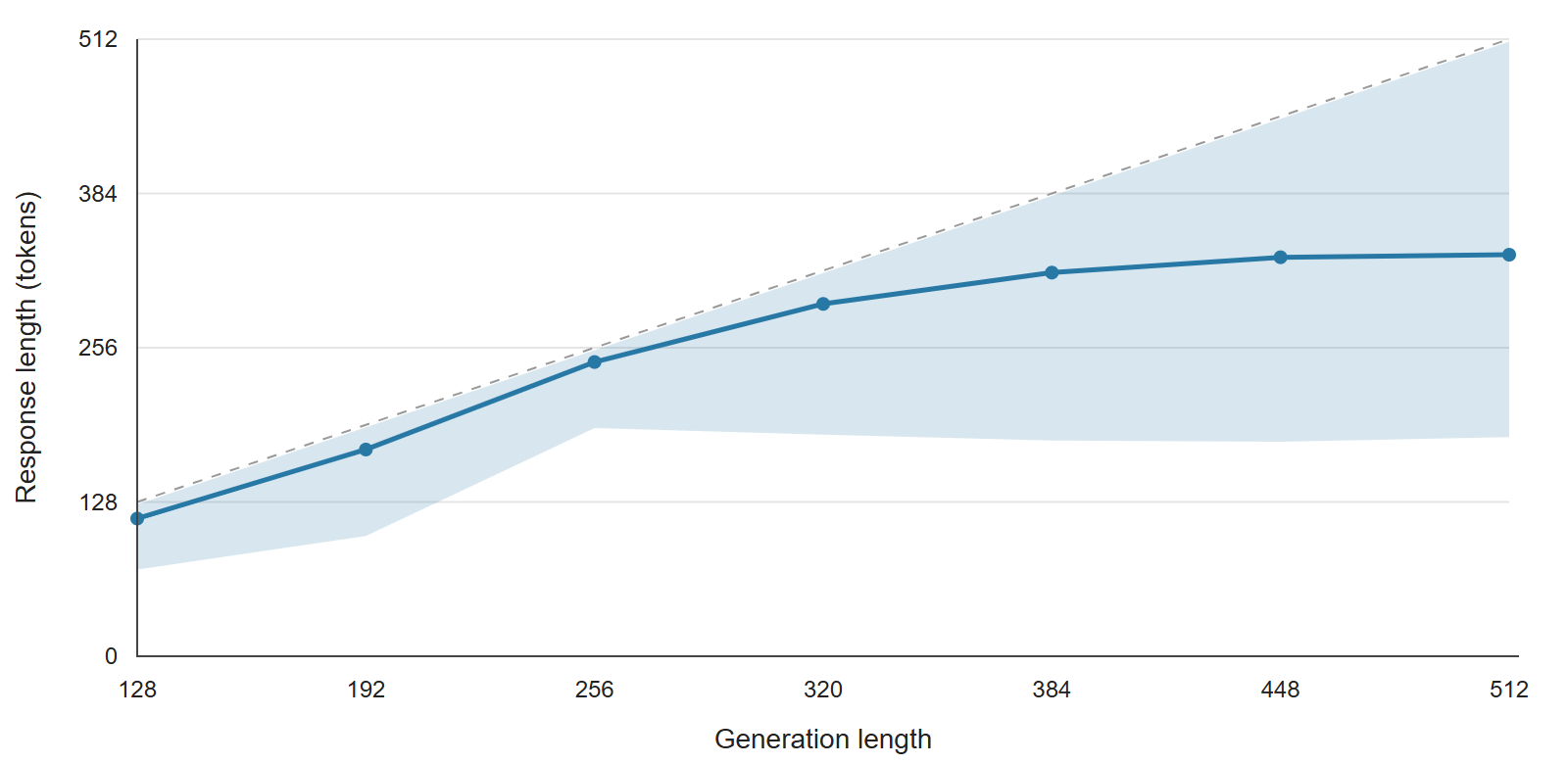}
    \caption{
    \textbf{Longer canvases induce longer responses.}
    Realized response length increases with allocated generation length.
    }
    \label{fig:res_len}
\end{figure}

First, random masking can omit reference content that the model's
reveal process would establish before selecting a target, or expose
later content prematurely. These cases can respectively penalize
missing-context uncertainty or conceal errors present at token
selection (Figure~\ref{fig:intro}A).

Second, canvas extent can change reference predictability even when
the solution is fixed. In our setting, longer canvases produce longer
responses (Figure~\ref{fig:res_len}), and changing extent shifts
reference loss, particularly in later response blocks
(Figure~\ref{fig:blockce}). Supervision may therefore compensate for
differences in response formulation rather than reasoning errors;
Figure~\ref{fig:intro}B illustrates this at termination.

These observations motivate supervision at token selection, rather
than uniformly stronger or weaker supervision. We therefore develop
PACS, which selects a canvas using commitment-level reference
predictability along a confidence-guided, reference-conditioned
reveal process and fine-tunes on the corresponding pre-commitment
states, without changing the reference.

\begin{samepage}
Our contributions are as follows:
\begin{itemize}

    \item We identify two sources of how training-canvas choices can misdirect supervision in full-attention DLM:
    the visible reference content revealed when a target is supervised, and the attention-visible extent of the training canvas.
    Both shift the prediction problem induced by the same fixed
    reference.

    \item We introduce \emph{PACS}, a predictability-aware SFT procedure
    that addresses these two mismatches by selecting the initial canvas
    through commitment-level reference predictability and supervising the
    corresponding pre-commitment states along the resulting
    reference-conditioned trajectory.

    \item Across four reasoning and code-generation benchmarks,
    PACS achieves the highest average accuracy among the compared
    SFT methods on both LLaDA-8B-Instruct and Dream-7B-Instruct
    at every evaluated generation budget, with individual-benchmark gains of up to 12.84 percentage points over vanilla SFT.

\end{itemize}
\end{samepage}

\section{Preliminaries}
\label{sec:background}

\subsection{Masked Diffusion Language Models and Canvases}
\label{sec:masked_diffusion}

We consider full-attention masked DLMs that operate on a prompt
followed by a preallocated region of token positions.
This region can extend beyond the eventual response and include
attention-visible padding, such as positions added for batching.
At the start of decoding, all non-prompt positions contain
$\mathtt{[mask]}$ tokens; generation proceeds by iteratively revealing selected predictions.

We call the complete attention-visible input the \emph{canvas}.
For a tokenized prompt $P$, we write the canvas as
$C=(C_1,\ldots,C_L)$, where $L$ counts all attention-visible positions.
The non-prompt canvas positions are
$\mathcal I_L=\{|P|+1,\ldots,L\}$ and contain visible tokens or
$\mathtt{[mask]}$.
We write $p_{\theta,i}(\cdot\mid C)$ for the token distribution
at position $i$ produced by the model with parameters $\theta$.

For a prompt--response pair $(P,A)$, where $A$ includes its
terminating $\mathtt{EOS}$ token, the clean canvas has token values
\begin{equation}
(y_1,\ldots,y_L)
=
[P,A,\mathtt{EOS}^{\,L-|P|-|A|}],
\qquad L\geq |P|+|A|.
\label{eq:clean_canvas}
\end{equation}
The $\mathtt{EOS}$ ending $A$ marks the reference boundary;
subsequent $\mathtt{EOS}$ copies fill the remaining attention-visible positions. Termination and padding are thus different roles of the same token value.
Here $y_i$ specifies the clean target at each non-prompt position.

\subsection{Confidence-Guided Decoding}
\label{sec:confidence_decoding}

For a fixed canvas length $L$, let $C^{(s)}$ denote the canvas
before decoding step $s$, initialized as
$C^{(0)}=[P,\mathtt{[mask]}^{\,L-|P|}]$.
Full-canvas decoding makes all masked positions in $\mathcal I_L$
eligible for revelation.
Semi-AR decoding instead processes contiguous response
blocks from left to right, restricting eligibility to the first
unfinished block; future blocks remain attention-visible as masks.

At each eligible position $i$, confidence is defined as the
probability of the model's most likely token:
\begin{equation}
c_i(C)=\max_{v\in\mathcal V}p_{\theta,i}(v\mid C),
\label{eq:prediction_confidence}
\end{equation}
where $\mathcal V$ is the non-mask output vocabulary.
For a threshold $\tau\in(0,1)$, the commitment set $S^{(s)}$
contains eligible positions with $c_i(C^{(s)})\geq\tau$.
Their predicted tokens are revealed together to form $C^{(s+1)}$,
after which confidence is recomputed.

\subsection{Training Canvases in Supervised Fine-Tuning}
\label{sec:vanilla_sft}

During supervised fine-tuning (SFT), padding accommodates
variable-length prompt--response pairs in batched training.
Some recipes set $L$ to the longest prompt--response sequence
in each minibatch
\citep{hu2026residualcontextdiffusionlanguage,li2025lavida},
while others use a dataset-level or otherwise globally fixed
training extent
\citep{zhao2025d1,nara2026,parashar2026lift,
chen2026dparallel}.

Vanilla SFT samples a masking probability
$t\sim\mathrm{Unif}(0,1)$ and constructs a training canvas $C$
from the clean tokens in Equation~\ref{eq:clean_canvas}.
Each $y_i$ at a non-prompt position $i\in\mathcal I_L$ is
independently replaced by $\mathtt{[mask]}$ with probability $t$,
including the $\mathtt{EOS}$-padding tokens; the prompt is unchanged.
The objective is
\begin{equation}
\mathcal L_{\mathrm{SFT}}(\theta)
=
-\mathbb E_{(P,A),\,t,\,C}
\left[
\frac{1}{t(L-|P|)}
\sum_{i\in\mathcal I_L}
\mathbf 1\!\left[C_i=\mathtt{[mask]}\right]
\log p_{\theta,i}\!\left(y_i\mid C\right)
\right].
\label{eq:vanilla_sft}
\end{equation}
The expectation is over training pairs, masking probabilities,
and sampled training canvases.
Here, $\mathbf 1[\cdot]$ denotes an indicator. Under this objective, every masked padding position is supervised
to reconstruct $\mathtt{EOS}$ and contributes to the loss.
Both masked and unmasked padding positions remain attention-visible. Thus, padding determines the attention-visible extent, while random masking determines which token values are visible and
which positions receive supervision.

\paragraph{Reference-conditioned revelation.}
For our diagnostics and PACS, we use a reference-conditioned counterpart
of this decoding process. At each step $s$, model confidence on the current
canvas determines the selected positions $S^{(s)}$, but their masks are
replaced by the corresponding clean reference values from Eq.~(1),
rather than the model's predicted tokens:
\[
C_i^{(s+1)} =
\begin{cases}
y_i, & i \in S^{(s)},\\
C_i^{(s)}, & \text{otherwise}.
\end{cases}
\]
At reference positions, $y_i=A_{i-|P|}$; positions beyond the reference
use the EOS-padding values. Thus, the reference determines the revealed
token values, while model confidence determines when they become visible.
Confidence is recomputed after each update.

\section{Related Work}
\label{sec:related_work}
\paragraph{Supervised fine-tuning for diffusion language models.}
Vanilla masked-diffusion SFT reconstructs randomly masked reference
tokens \citep{sahoo2024simple,shi2024simplified,nie2025large}.
Because masks are sampled independently of model predictions, training
contexts need not reflect confidence-guided decoding.
Recent methods incorporate decoding structure or token predictability.
Block-based approaches condition prediction on preceding clean blocks
\citep{arriola2025block,chen2026dparallel,jain2026adaptive}. \citet{chen2026trims} guide masking with teacher-derived difficulty
rankings, while \citet{parashar2026lift} adapt target selection to
model-estimated difficulty and visible context.
PACS shares this motivation of using model predictability to shape
supervision, but follows confidence-guided decoding more directly:
after each reference revelation, confidence is recomputed to determine
which positions are supervised next.

\paragraph{Trajectory-aware training and reference supervision.}
Recent work brings training closer to confidence-guided inference by
learning from states visited during the model's own decoding process
\citep{wang2026tracerl,shao2026back,ren2026topd}.
For on-policy reinforcement learning and distillation, these trajectories
can be followed directly because model-generated tokens remain part of
subsequent states.
Fixed-reference SFT cannot follow such trajectories exactly: teacher
forcing replaces model predictions with reference tokens, changing the
states from which later decisions are made.
PACS therefore takes a middle ground, retaining reference supervision
while letting model predictability guide the evolving reveal process. Plan-Verify-Fill (PVF)~\citep{li2026plan} also builds on
confidence-guided decoding, but intervenes at inference time by drafting
and verifying additional token predictions without updating model weights.
In contrast, PACS uses confidence-guided reveal order to construct
reference-conditioned supervision during fine-tuning.

\paragraph{From response compatibility to canvas compatibility.}
Recent SFT methods adapt supervision to the learner's response
distribution through likelihood-based response selection
\citep{zhang2026best}, reference rewriting
\citep{yang2024self}, or selective use of correct model-generated
answers \citep{gupta2025selective}.
These approaches seek to teach the task while limiting unnecessary
changes to the model's existing response patterns.
PACS extends this principle to a degree of freedom absent from
standard AR teacher forcing: \emph{the attention-visible
extent of the training canvas}.
In full-attention masked DLMs, changing this extent can alter
reference predictability even when the solution itself is unchanged.
Compatibility therefore depends not only on the reference response,
but also on the canvas on which it is taught.
PACS preserves the reference and selects its canvas using
commitment-level reference predictability, adapting the prediction
conditions rather than the answer.

\section{How Canvas Shapes Reference Supervision}
\label{sec:conditioning_analysis}
\begin{figure}[t]
    \centering    \includegraphics[width=0.85\linewidth]{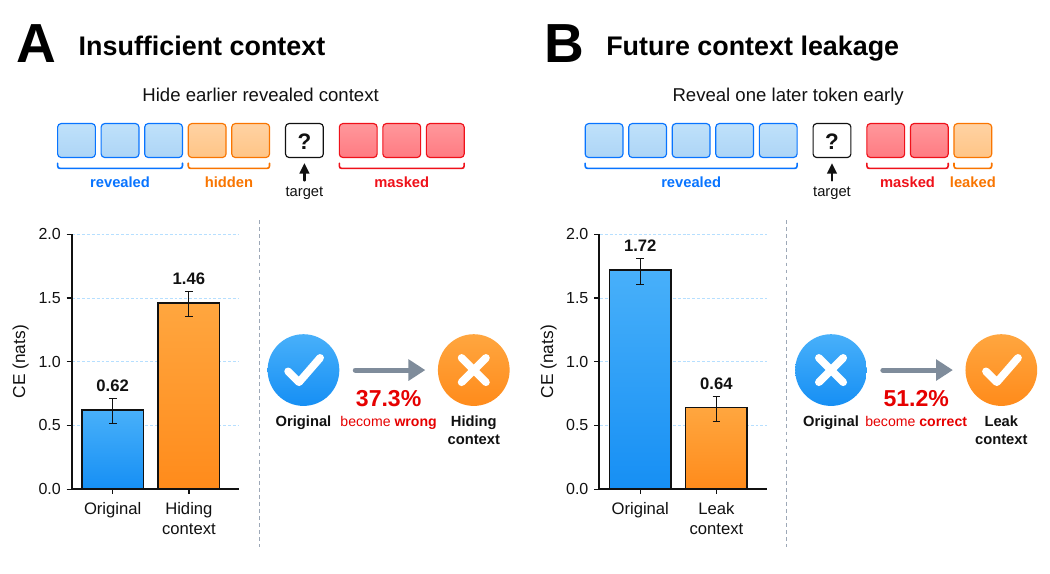}
    \caption{
    \textbf{Changing context around commitment can create or conceal
    reference disagreement.}
    LLaDA-8B-Instruct on GSM8K.
    \textbf{A: Insufficient context.}
    Hiding reference tokens revealed in the preceding step makes
    37.3\% of initially correct predictions wrong.
    \textbf{B: Future context leakage.}
    Revealing one reference token scheduled for the following reveal step
    early makes 51.2\% of initially wrong predictions correct.}
     \label{fig:context_interventions}\end{figure}

In this section, we examine the canvas mismatches that motivate PACS.
We show how changing the reference content available at supervision
or the canvas length changes predictions for the same
reference response.
These observations motivate when PACS supervises each target and
how it selects the training canvas.
All diagnostics use a frozen model; experimental details appear
in Appendix~\ref{app:sec3exp}.

\subsection{The Canvas Can Create or Conceal Commitment Disagreement}
\label{sec:uncertainty_before_commitment}

We construct reference-conditioned trajectories by selecting
masked positions according to model confidence and revealing
their reference tokens before the next selection.
A target's prediction immediately before revelation is its
\emph{commitment prediction}.
We call it \emph{correct} (reference-agreeing) when the most likely token matches the reference, and \emph{wrong} otherwise.
Starting from this original canvas, we change the surrounding content
while keeping the target masked and the canvas length fixed.
Figure~\ref{fig:context_interventions} compares mean reference CE
before and after each change and reports the corresponding
correct-to-wrong or wrong-to-correct transitions.

\paragraph{Insufficient context.}
We consider targets predicted correctly at commitment that were
already eligible in the immediately preceding reveal step.
For each target, we hide the reference tokens revealed in that step
by replacing them with $\mathtt{[mask]}$, leaving the target masked
and all other positions unchanged.
After hiding this context, 37.3\% of these initially correct predictions
become wrong, and mean reference CE increases by 0.84 nats
(Figure~\ref{fig:context_interventions}A).
The additional supervision thus penalizes uncertainty from
missing context, rather than an error at the target's original
selection step.

\paragraph{Future context leakage.}
We instead consider targets predicted incorrectly at commitment.
Starting from each target's original canvas, we reveal one reference
token selected in the following reveal step of the original trajectory,
leaving the target masked and all other positions unchanged.
Revealing this token early makes 51.2\% of these initially wrong
predictions correct and reduces mean reference CE by 1.08 nats
(Figure~\ref{fig:context_interventions}B).
Conversely, future context leakage can weaken corrective supervision
at positions the model still predicts incorrectly at commitment.
By making these targets appear easier, the leaked information suppresses
the reference-loss signal for errors that remain on the original canvas.

Together, these results show that canvas content can misdirect corrective supervision relative
to the model's errors at commitment.
These findings motivate supervising each target on the canvas present
when the confidence-guided rule selects it for revelation.

\subsection{Canvas Extent Changes Reference Predictability}
\label{sec:canvas_length_predictability}

Even when the reference response is unchanged, varying the canvas
length can change the probabilities assigned to its tokens. Section~\ref{sec:uncertainty_before_commitment} motivates commitment-level
scoring as a way to assess how well a canvas supports the reference when
its tokens are selected for revelation.
We therefore use mean CE at commitment to examine how this
predictability changes with canvas extent.
Let $C^{(s)}$ be the canvas before reveal step $s$,
$S_L^{(s)}$ the positions selected by the confidence-guided rule,
and $T_L$ the total number of reveal steps:
\begin{equation}
\mathrm{CE}_{\mathrm{tr}}(P,A;L)
=
-\frac{1}{|\mathcal{R}|}
\sum_{s=0}^{T_L-1}
\sum_{i\in S_L^{(s)}\cap\mathcal R}
\log p_{\theta,i}\!\left(y_i\mid C^{(s)}\right).
\label{eq:trace_tight_score}
\end{equation}
Here, $\mathcal R$ contains the reference positions through the first
response-termination token, inclusive, and $y_i$
is the reference token at position $i$.
Each target is scored once, immediately before its revelation.

To study this effect, we generate references at source generation length $K_0=384$ and
reevaluate them at other lengths, where $K=L-|P|$.
Each candidate uses its own reference-conditioned trajectory, and CE changes are paired against the same reference evaluated at $K_0$.

\begin{figure}[t]
    \centering
\includegraphics[width=\linewidth]{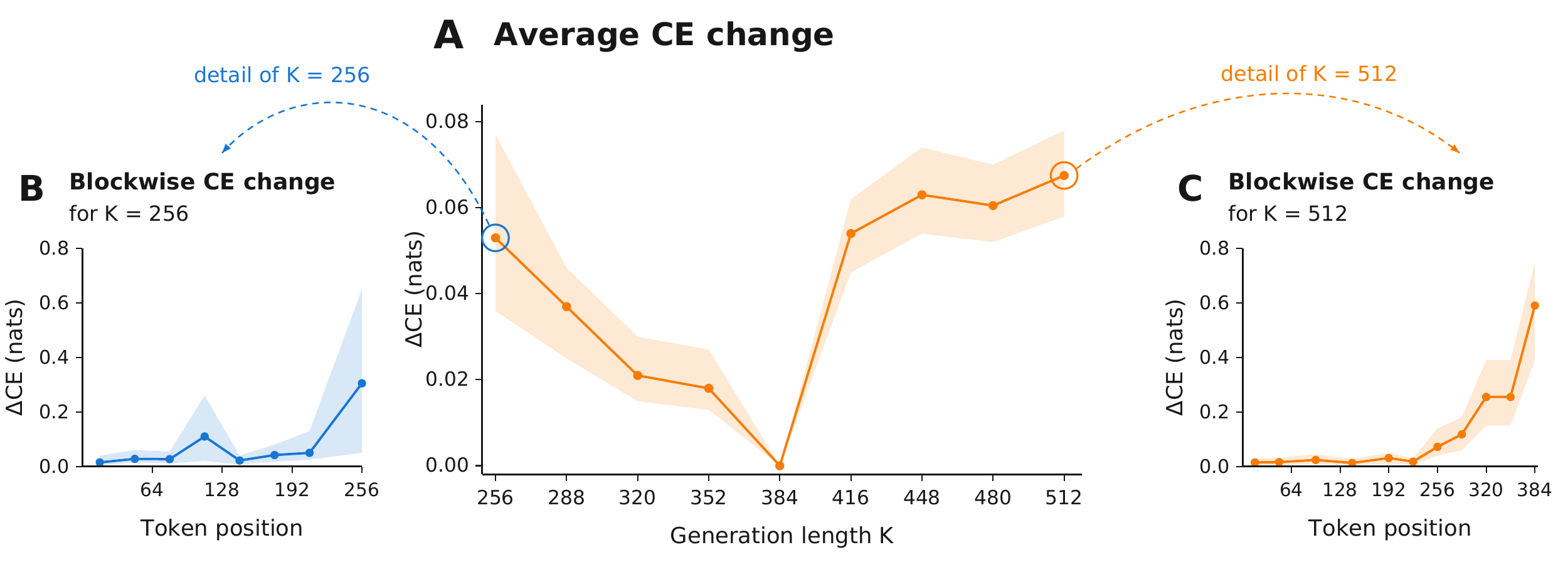}
    \caption{
    \textbf{Canvas-extent mismatch increases reference cross-entropy.}
Target responses are generated at source generation length $K_0=384$, and $\Delta\mathrm{CE}$ is the change of CE compared to that evaluated at $K_0$.
\textbf{A (center):} Mean $\Delta\mathrm{CE}$ across generation lengths,
evaluated along each candidate's own reference-conditioned trajectory.
\textbf{B (left) and C (right):} Detailed blockwise $\Delta\mathrm{CE}$ across
response positions for $K=256$ (blue) and $K=512$ (orange).}
    \label{fig:blockce}
\end{figure}

Figure~\ref{fig:blockce}A shows positive mean paired
$\Delta\mathrm{CE}$ at both shorter and longer lengths.
Panels B and C show that the excess CE is particularly pronounced
in later response blocks for $K=256$ and $K=512$. This additional loss need not reflect a reasoning error.
It may also reflect differences between the model's response
behavior under a given canvas and the formulation of the fixed
reference. The termination example in Figure~\ref{fig:intro}B illustrates one such discrepancy.
For these source-generated references, the observed sensitivity
to extent motivates using $\mathrm{CE}_{\mathrm{tr}}$ as an
empirical measure of compatibility between a fixed reference
and a candidate canvas.

\section{Predictability-Aware Canvas Selection and Supervision}
\label{sec:method}

\begin{algorithm}[t]
\caption{PACS: canvas selection and commitment-level SFT}
\label{alg:sec5}
\begin{algorithmic}[1]
\Require Training pairs $\mathcal D$, initial model $p_{\theta_0}$,
    candidate lengths $\Lambda(P,A)$
\State Initialize $\theta\gets\theta_0$
\ForAll{current training pairs $(P,A)\in\mathcal D$}
        \State Construct $\mathcal T_L(P,A)$ for each
            $L\in\Lambda(P,A)$ using current $p_\theta$
            with confidence-guided reference revelation
        \State Score each candidate with
            $\mathrm{CE}_{\mathrm{tr}}(P,A;L)$
        \State Select $L^*(P,A)$ by
            \eqref{eq:selected_training_length}
        \State Fine-tune $p_\theta$ using
        \eqref{eq:commitment_sft_objective} on the retained pre-commitment canvases,
    supervising only selected reference positions
\EndFor
\State \Return $p_\theta$
\end{algorithmic}
\vspace{-2pt}
\end{algorithm}
Section~\ref{sec:conditioning_analysis} shows that both canvas content and
canvas extent shape reference supervision.
Accordingly, PACS has two components:
\emph{commitment-based canvas selection}, which chooses a canvas under which the
reference is well supported along the model's reveal trajectory, and
\emph{commitment-level SFT}, which teaches the reference at the
corresponding pre-commitment states.

\subsection{Commitment-Based Canvas Selection}
\label{sec:canvas_selection}

Recall that the clean canvas extends the reference with
attention-visible $\mathtt{EOS}$ padding.
When replaced by $\mathtt{[mask]}$ and included in supervision,
these positions are trained to reconstruct $\mathtt{EOS}$ and
contribute to the loss.
PACS selects this attention-visible canvas extent for each fixed
reference using commitment-level predictability, rather than
inheriting it from a padding convention.

Let $\mathcal D$ contain the training pairs $(P,A)$.
For each pair, let $\mathcal R$ be the scored reference positions
defined in Section~\ref{sec:canvas_length_predictability}, and let
$\Lambda(P,A)$ be a finite, nonempty set of candidate total canvas
lengths satisfying $L\geq |P|+|A|$.
Each candidate retains the same prompt and reference but allocates
$L-|P|-|A|$ attention-visible positions beyond the reference.

For each $L\in\Lambda(P,A)$, the model $p_{\theta}$ constructs
the reference-conditioned trajectory from Section~\ref{sec:uncertainty_before_commitment}
(details in Appendix~\ref{app:sec3exp}).
The padding positions begin as $\mathtt{[mask]}$ and take their clean
$\mathtt{EOS}$ values when revealed.
We record the pre-commitment canvases and commitment sets as
\begin{equation}
    \mathcal T_L(P,A)
    =
    \bigl(C^{(s)},S_L^{(s)}\bigr)_{s=0}^{T_L-1},
    \label{eq:recorded_training_trace}
\end{equation} with every position in $\mathcal R$ committed once along the trajectory.

We compare candidates using
$\mathrm{CE}_{\mathrm{tr}}(P,A;L)$ from
Equation~\ref{eq:trace_tight_score}, scoring only the same reference
span $\mathcal R$ across all lengths.
Thus, the additional padding affects the conditioning canvas but
does not add targets to the selection score.
We select
\begin{equation}
L^*(P,A)
\in
\operatorname*{arg\,min}_{L\in\Lambda(P,A)}
\mathrm{CE}_{\mathrm{tr}}(P,A;L),
\label{eq:selected_training_length}
\end{equation}
and retain the selected canvas length and trajectory
$\mathcal T_{L^*}(P,A)$ for SFT in
Section~\ref{sec:commitment_sft}.

\subsection{Commitment-Level Supervised Fine-Tuning}
\label{sec:commitment_sft}

For each training pair, let $L^*=L^*(P,A)$.
The selected trajectory $\mathcal T_{L^*}(P,A)$ supplies the
pre-commitment canvases for fine-tuning.
Rather than applying loss to all unresolved positions on each canvas,
we supervise only the positions selected for commitment:
\begin{equation}
\mathcal L_{\mathrm{commit}}(\theta)
=
-\mathbb E_{(P,A)\sim\mathcal D}
\left[
\frac{1}{|\mathcal R|}
\sum_{s=0}^{T_{L^*}-1}
\sum_{i\in S_{L^*}^{(s)}\cap\mathcal R}
\log p_{\theta,i}\!\left(y_i\mid C^{(s)}\right)
\right].
\label{eq:commitment_sft_objective}
\end{equation}
Each reference position in $\mathcal R$ is therefore supervised once along
the selected trajectory.

This implements \emph{confidence-guided teacher forcing}: the model
determines which positions are selected at each step, while the reference
supplies the values revealed before the next prediction.
The resulting canvases remain reference-consistent, but their reveal order
follows the model's own confidence-guided decoding behavior. Supervising the pre-commitment prediction focuses the loss on the prediction problem encountered when the decoder elects to commit.
The method therefore changes the context in which each reference token is
taught, rather than removing difficult targets.

Canvas selection and commitment-level SFT operate on the same conditional
prediction tasks: candidate canvases are assessed through their
pre-commitment reference predictions, and the selected trajectory supplies
those same conditioning states for fine-tuning.
Integrating these two components,
Algorithm~\ref{alg:sec5} summarizes the complete
decoding-aware SFT procedure.

\section{Experiments}
\label{sec:experiments}
We evaluate downstream generation, isolate the contribution of each PACS
component to examine how canvas selection changes reference supervision.

\subsection{Experimental Setup}
\label{sec:experimental_setup}
We evaluate LLaDA-8B-Instruct and Dream-7B-Instruct on GSM8K, MATH500,
MBPP, and HumanEval.  Each checkpoint is evaluated at $K_{\mathrm{eval}}\in\{128,256,512\}$. Tables~\ref{tab:main_results}
and~\ref{tab:dream_results} show the $K_{\mathrm{eval}}=256$ results;
bold marks the highest score in each benchmark column. Complete
three-length scores and the relative percentage gains $\Delta$ of PACS
over the corresponding initial checkpoint appear in
Tables~\ref{tab:full_llada_results} and~\ref{tab:full_dream_results}. Additional training
details are provided in Section~\ref{app:exp_setup}.

\paragraph{Data and comparisons.}
\label{sec:models_training_data}
\label{sec:baselines_controls}
For a fair comparison, all SFT methods use the same training data,
prepared from the training splits of GSM8K
\citep{cobbe2021gsm8k} and MATH \citep{hendrycks2021math}
and from OpenCoder \citep{huang2025opencoder}.
We follow the reference-guided response-construction procedure
of ReGFT \citep{wu2026regft}, using confidence-guided decoding
\citep{wu2025fastdllm} to generate model-written reference responses.
We compare vanilla random-mask SFT~\citep{sahoo2024simple,shi2024simplified,nie2025large}, semi-AR block SFT \citep{jain2026adaptive,chen2026dparallel}, LIFT~\citep{parashar2026lift}, TRIMS~\citep{chen2026trims}, and
PACS.

\paragraph{Canvas search.}
For each pair $(P,A)$, let $L_{\mathrm{ref}}=|P|+|A|$ be the
minimum feasible total canvas length and
$L_{\min}=B\lceil L_{\mathrm{ref}}/B\rceil$ its rounded value,
with $B=32$. We consider the nested families
\[
\Lambda_K(P,A)=
\begin{cases}
\{L_{\mathrm{ref}}\}, & K=1,\\[2pt]
\{L_{\mathrm{ref}}\}
\cup \{L_{\min}+jB : j=0,\ldots,K-2\},
& K=2,\ldots,5.
\end{cases}
\]
Duplicate lengths are evaluated once. Our main family is
$\Lambda_{\mathrm{main}}=\Lambda_4$. PACS selects each example's
canvas using commitment-level reference CE
(Section~\ref{sec:canvas_selection}); the search extent is chosen
on held-out validation (Section~\ref{sec:pacs-extent}).
\paragraph{Main results.}
PACS outperforms vanilla SFT in 22 of 24 model--benchmark--length
settings and ties in the remaining two, achieving the highest
accuracy among all compared methods in 19 settings (Tables~\ref{tab:full_llada_results} and~\ref{tab:full_dream_results}).
At $K_{\mathrm{eval}}=256$, it improves MATH500 by 2.00 and 5.00
percentage points and HumanEval by 9.15 and 6.71 points on LLaDA
and Dream, respectively (Tables~\ref{tab:main_results}
and~\ref{tab:dream_results}). The full results show that these gains
extend across both backbones and all three generation budgets.

\begin{table*}[t]
    \centering
    \caption{LLaDA-8B-Instruct results (\%) at $K_{\mathrm{eval}}=256$.}
    \label{tab:main_results}
    \begingroup
    \normalsize
    \setlength{\tabcolsep}{4pt}
    \renewcommand{\arraystretch}{1.1}
    \begin{tabular*}{\textwidth}{@{\extracolsep{\fill}}lrrrr@{}}
        \toprule
        Method & GSM8K & MATH500 & MBPP & HumanEval \\
        \midrule
        Initial & 75.44 & 36.40 & 40.86 & 37.80 \\
        Vanilla SFT & 76.27 & 37.00 & 49.03 & 38.41 \\
        Semi-AR block & \textbf{78.20} & 38.40 & 48.60 & 43.90 \\
        LIFT & 77.63 & 37.00 & 46.69 & 39.63 \\
        TRIMS & 77.63 & 38.00 & \textbf{52.92} & 40.24 \\
        \midrule
        PACS & 77.94 & \textbf{39.00} & 49.42 & \textbf{47.56} \\
        \bottomrule
    \end{tabular*}
    \endgroup
\end{table*}

\begin{table*}[t]
    \centering
    \caption{Dream-7B-Instruct results (\%) at $K_{\mathrm{eval}}=256$.}
    \label{tab:dream_results}
    \begingroup
    \normalsize
    \setlength{\tabcolsep}{4pt}
    \renewcommand{\arraystretch}{1.1}
    \begin{tabular*}{\textwidth}{@{\extracolsep{\fill}}lrrrr@{}}
        \toprule
        Method & GSM8K & MATH500 & MBPP & HumanEval \\
        \midrule
        Initial & \textbf{80.52} & 43.80 & 45.53 & 51.83 \\
        Vanilla SFT & 80.06 & 44.20 & 60.31 & 54.27 \\
        Semi-AR block & 80.21 & 43.60 & 61.09 & 58.53 \\
        LIFT & 80.36 & 44.80 & 50.58 & 52.44 \\
        TRIMS & 80.44 & 45.60 & 52.14 & 53.66 \\
        \midrule
        PACS & \textbf{80.52} & \textbf{49.20} & \textbf{61.48} & \textbf{60.98} \\
        \bottomrule
    \end{tabular*}
    \endgroup
\end{table*}

\subsection{Component Analysis}
\label{sec:component_analysis}
\label{sec:joint_component_ablation}
We cross two canvas rules with random or confidence-guided ordering.
Within each canvas rule, random ordering matches the corresponding
confidence trajectory's per-step reveal counts. Appendix~\ref{app:random_order} details the schedule-matched control.

Table \ref{tab:component_study}  shows that canvas selection improves both reveal-order variants, while confidence-guided
ordering outperforms schedule-matched random ordering under either canvas rule. Combining both
components achieves the highest accuracy on all four benchmarks, supporting complementary benefits from selecting the canvas extent and the pre-commitment states used for supervision.

\begin{table*}[!ht]
    \centering
    \caption{{Contributions of canvas selection and reveal ordering.}
    LLaDA at $K_{\mathrm{eval}}=256$.}
    \label{tab:component_study}
    \begingroup
    \normalsize
    \setlength{\tabcolsep}{2pt}
    \renewcommand{\arraystretch}{1.35}
    \begin{tabular*}{\textwidth}{@{\extracolsep{\fill}}llrrrr@{}}
        \toprule
        Canvas rule & Reveal order & GSM8K & MATH500 & MBPP & HumanEval \\
        \midrule
        Batch-max & Random
        & 76.27 & 37.00 & 45.14 & 39.63 \\
        Batch-max & Confidence-guided (ours)
        & 77.48 & 38.00 & 47.08 & 40.90 \\
        \midrule
        Trajectory-selected (ours) & Random
        & 77.03 & 37.40 & 47.86 & 44.51 \\
        Trajectory-selected (ours) & Confidence-guided (ours)
        & 77.94 & 39.00 & 49.42 & 47.56 \\
        \bottomrule
    \end{tabular*}
    \endgroup
\end{table*}
\subsection{Canvas Choice and Reference Predictability}
\label{sec:canvas_choice_predictability}

\begin{figure*}[!ht]
    \centering
    \includegraphics[width=0.95\textwidth]
{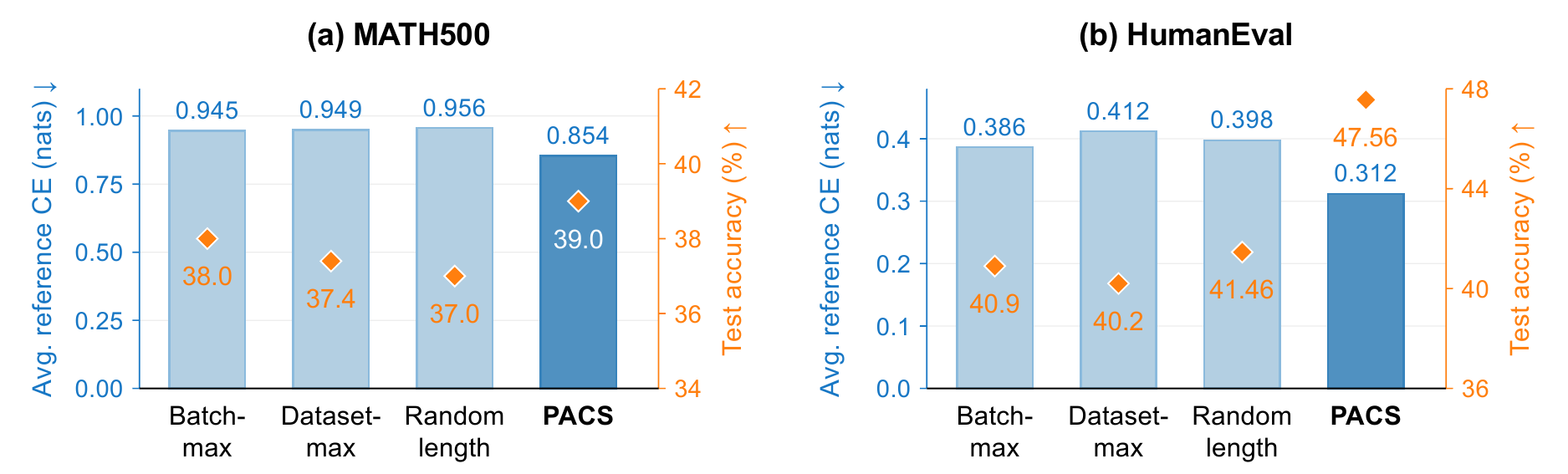}
    \caption{\textbf{Canvas choice and reference predictability.}
    Target CE (blue bars, left) and post-SFT test accuracy
    (orange diamonds, right) for four canvas-length rules on MATH500
    and HumanEval with LLaDA-8B-Instruct at $K_{\mathrm{eval}}=256$.}
    \label{fig:canvas_choice}
\end{figure*}

We compare predictability-based canvas selection with length rules
motivated by existing SFT recipes. Batch-max uses the longest example in each batch, representing batch-dependent padding. Dataset-max uses a shared maximum length across training examples,
providing a control for globally specified training extents.
We additionally include Random visible length, which samples
feasible canvas lengths without predictability-based selection
\citep{ye2025dream}. Figure~\ref{fig:canvas_choice} compares target CE
and downstream test performance after SFT on MATH500 and HumanEval
at $K_{\mathrm{eval}}=256$.

With the reference responses unchanged, PACS yields lower reference CE
and higher test accuracy than batch-max, dataset-max, and random
visible lengths (Figure~\ref{fig:canvas_choice}).

\subsection{Ablation on canvas search extent}
\label{sec:pacs-extent}

We vary the nested candidate family $\Lambda_K$ defined in
Section~\ref{sec:experimental_setup}, holding the
commitment-level reference CE selection criterion fixed.

\begin{figure}[htbp]
    \centering
\includegraphics[width=0.95\linewidth]{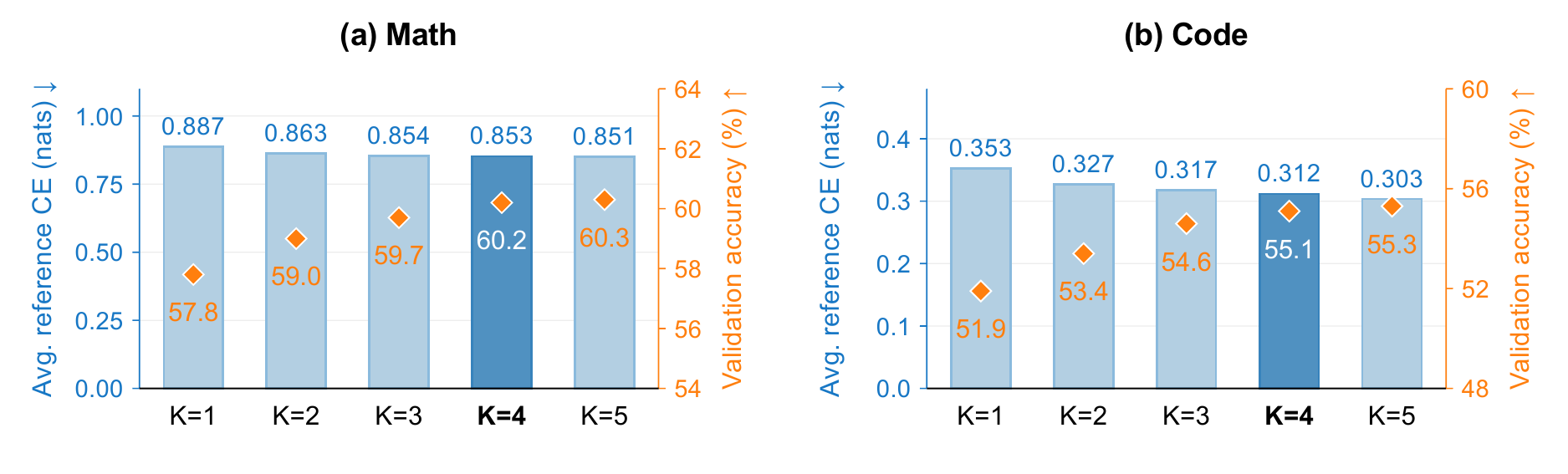}
    \caption{
        Canvas search extent at $K_{\mathrm{eval}}=256$.
        Bars show average reference CE (left axis);
        orange diamonds show held-out validation accuracy (right axis).}
    \label{fig:pacs-search-extent}
\end{figure}

Most validation improvement is obtained by our main family
$\Lambda_4$; expanding to $\Lambda_5$ adds only 0.1 percentage
points on Math and 0.2 points on Code
(Figure~\ref{fig:pacs-search-extent}). A compact candidate family
therefore captures most of the observed benefit while limiting
per-example search.

\section{Conclusion}
This work highlights training-canvas construction as an integral part
of masked-DLM fine-tuning: revealed reference content and
attention-visible extent shape the learning problem even when the
answer is fixed. PACS couples canvas selection and reference-consistent
pre-commitment supervision through commitment-level reference
predictability, bringing the model's decoding behavior into the
construction of its training tasks. Experiments across mathematical
reasoning, code generation, and multiple generation lengths support
this decoding-aware perspective on supervision. More broadly, this work opens promising directions for future research
on predictability-aware canvas construction across diverse DLM
architectures, including block diffusion and hybrid
diffusion--AR models.

\subsection*{AI use statement}
AI assistance with manuscript writing was limited to language polishing, including improvements to grammar, clarity, and readability. The authors reviewed all AI-assisted revisions and take full responsibility for the final manuscript.

\bibliography{main}

@inproceedings{lou2024discrete,
  title = 	 {Discrete Diffusion Modeling by Estimating the Ratios of the Data Distribution},
  author =       {Lou, Aaron and Meng, Chenlin and Ermon, Stefano},
  booktitle = 	 {Proceedings of the 41st International Conference on Machine Learning},
  pages = 	 {32819--32848},
  year = 	 {2024},
  volume = 	 {235},
  series = 	 {Proceedings of Machine Learning Research},
  month = 	 {21--27 Jul},
  publisher =    {PMLR}
}

@inproceedings{li2026plan,
  title={Plan, verify and fill: A structured parallel decoding approach for diffusion language models},
  author={Li, Miao and Jiang, Hanyang and Cheng, Sikai and Fu, Hengyu and Cai, Yuhang and Huang, Baihe and Chen, Zhongzhu and Ye, Tinghan and Chen, Xuanzhou and Van Hentenryck, Pascal},
  booktitle={Findings of the Association for Computational Linguistics: EMNLP 2026},
  pages={23694--23717},
  year={2026}
}

@inproceedings{yang2024self,
  title={Self-distillation bridges distribution gap in language model fine-tuning},
  author={Yang, Zhaorui and Pang, Tianyu and Feng, Haozhe and Wang, Han and Chen, Wei and Zhu, Minfeng and Liu, Qian},
  booktitle={Proceedings of the 62nd Annual Meeting of the Association for Computational Linguistics (Volume 1: Long Papers)},
  pages={1028--1043},
  year={2024}
}

@inproceedings{gupta2025selective,
  title={Selective self-to-supervised fine-tuning for generalization in large language models},
  author={Gupta, Sonam and Nandwani, Yatin and Yehudai, Asaf and Khandelwal, Dinesh and Raghu, Dinesh and Joshi, Sachindra},
  booktitle={Findings of the Association for Computational Linguistics: NAACL 2025},
  pages={6255--6264},
  year={2025}
}

@inproceedings{sahoo2024simple,
  title     = {Simple and Effective Masked Diffusion Language Models},
  author    = {Sahoo, Subham Sekhar and
               Arriola, Marianne and
               Schiff, Yair and
               Gokaslan, Aaron and
               Marroquin, Edgar and
               Chiu, Justin T. and
               Rush, Alexander and
               Kuleshov, Volodymyr},
  booktitle = {Advances in Neural Information Processing Systems},
  volume    = {37},
  pages     = {130136--130184},
  year      = {2024},
  doi       = {10.52202/079017-4135},
  url       = {https://proceedings.neurips.cc/paper_files/paper/2024/hash/eb0b13cc515724ab8015bc978fdde0ad-Abstract-Conference.html}
}

@article{shao2026back,
  title={Back on Track: Aligning Rewards and States for Reasoning in Diffusion Large Language Models},
  author={Shao, Yawen and Xiao, Jie and Zhu, Kai and Liu, Yu and Luo, Hongchen and Fu, Xueyang and Cao, Yang and Zhai, Wei and Zha, Zheng-Jun},
  journal={arXiv preprint arXiv:2606.08501},
  year={2026}
}

@inproceedings{shi2024simplified,
  title     = {Simplified and Generalized Masked Diffusion
               for Discrete Data},
  author    = {Shi, Jiaxin and
               Han, Kehang and
               Wang, Zhe and
               Doucet, Arnaud and
               Titsias, Michalis K.},
  booktitle = {Advances in Neural Information Processing Systems},
  volume    = {37},
  pages     = {103131--103167},
  year      = {2024},
  doi       = {10.52202/079017-3277},
  url       = {https://proceedings.neurips.cc/paper_files/paper/2024/hash/bad233b9849f019aead5e5cc60cef70f-Abstract-Conference.html}
}

@inproceedings{nie2025large,
  title     = {Large Language Diffusion Models},
  author    = {Nie, Shen and
               Zhu, Fengqi and
               You, Zebin and
               Zhang, Xiaolu and
               Ou, Jingyang and
               Hu, Jun and
               Zhou, Jun and
               Lin, Yankai and
               Wen, Ji-Rong and
               Li, Chongxuan},
  booktitle = {Advances in Neural Information Processing Systems},
  volume    = {38},
  year      = {2025},
  note      = {Oral presentation},
  doi       = {10.52202/085713-1689},
}

@article{xu2025kodcode,
  title={KodCode: A Diverse, Challenging, and Verifiable Synthetic Dataset for Coding},
  author={Xu, Zhangchen and Liu, Yang and Yin, Yueqin and Zhou, Mingyuan and Poovendran, Radha},
  journal={arXiv preprint arXiv:2503.02951},
  year={2025}
}

@inproceedings{wu2025fastdllm,
 title={Fast-dllm: Training-free acceleration of diffusion llm by enabling kv cache and parallel decoding},
  author={Wu, Chengyue and Zhang, Hao and Xue, Shuchen and Liu, Zhijian and Diao, Shizhe and Zhu, Ligeng and Luo, Ping and Han, Song and Xie, Enze},
  booktitle={International Conference on Learning Representations},
  volume={2026},
  pages={57027--57051},
  year={2026}
}

@inproceedings{arriola2025block,
  title={Block diffusion: Interpolating between autoregressive and diffusion language models},
  author={Arriola, Marianne and Gokaslan, Aaron and Chiu, Justin and Yang, Zhihan and Qi, Zhixuan and Han, Jiaqi and Sahoo, Subham and Kuleshov, Volodymyr},
  booktitle={International Conference on Learning Representations},
  volume={2025},
  pages={50726--50753},
  year={2025}
}

@article{zhang2026best,
  title={The best instruction-tuning data are those that fit},
  author={Zhang, Dylan and Dai, Qirun and Peng, Hao},
  journal={Advances in Neural Information Processing Systems},
  volume={38},
  pages={141172--141208},
  year={2026}
}

@misc{jain2026adaptive,
  title = {Adaptive Block Diffusion: Resolving Training-Inference
           Mismatch in Diffusion Language Models},
  author = {Gagan Jain},
  year = {2026},
  eprint = {2606.29275},
  archivePrefix = {arXiv},
  primaryClass = {cs.LG}
}

@inproceedings{wang2026tracerl,
  title={Revolutionizing reinforcement learning framework for diffusion large language models},
  author={Wang, Yinjie and Yang, Ling and Li, Bowen and Tian, Ye and Shen, Ke and Wang, Mengdi},
  booktitle={International Conference on Learning Representations},
  volume={2026},
  pages={101476--101501},
  year={2026}
}

@article{ren2026topd,
  title         = {Trace-Based On-Policy Distillation
                   for Masked Diffusion Language Models},
  author        = {Ren, Haolin and
                   Huang, Ziyang and
                   Yuan, Chenhao and
                   Zhao, Jun and
                   Liu, Kang},
  journal       = {arXiv preprint arXiv:2607.16872},
  year          = {2026},
  eprint        = {2607.16872},
  archivePrefix = {arXiv},
  primaryClass  = {cs.CL},
  url           = {https://arxiv.org/abs/2607.16872}
}

@inproceedings{chen2026dparallel,
  title         = {{dParallel}: Learnable Parallel Decoding for {dLLMs}},
  author        = {Chen, Zigeng and
                   Fang, Gongfan and
                   Ma, Xinyin and
                   Yu, Ruonan and
                   Wang, Xinchao},
  booktitle     = {The Fourteenth International Conference
                   on Learning Representations},
  year          = {2026},
  eprint        = {2509.26488},
  archivePrefix = {arXiv},
  primaryClass  = {cs.CL},
  url           = {https://proceedings.iclr.cc/paper_files/paper/2026/hash/57250222014c35949476f3f272c322d2-Abstract-Conference.html}
}

@article{ye2025dream,
  title         = {{Dream 7B}: Diffusion Large Language Models},
  author        = {Ye, Jiacheng and
                   Xie, Zhihui and
                   Zheng, Lin and
                   Gao, Jiahui and
                   Wu, Zirui and
                   Jiang, Xin and
                   Li, Zhenguo and
                   Kong, Lingpeng},
  journal       = {arXiv preprint arXiv:2508.15487},
  year          = {2025},
  eprint        = {2508.15487},
  archivePrefix = {arXiv},
  primaryClass  = {cs.CL},
  url           = {https://arxiv.org/abs/2508.15487}
}

@article{chen2026trims,
  title         = {{TRIMS}: Trajectory-Ranked Instruction Masked
                   Supervision for Diffusion Language Models},
  author        = {Chen, Lingjie and
                   Qiu, Ruizhong and
                   Fan, Yuyu and
                   Zhao, Yanjun and
                   Tong, Hanghang},
  journal       = {arXiv preprint arXiv:2604.00666},
  year          = {2026},
  eprint        = {2604.00666},
  archivePrefix = {arXiv},
  primaryClass  = {cs.CL},
  url           = {https://arxiv.org/abs/2604.00666}
}

@inproceedings{parashar2026lift,
title={Learnability-Informed Fine-Tuning of Diffusion Language Models},
author={Shubham Parashar and Atharv Chagi and Jacob Helwig and Lakshmi Jotsna Madhavarapu and Sushil Vemuri and James Caverlee and Dileep Kalathil and Shuiwang Ji},
booktitle={Forty-third International Conference on Machine Learning},
year={2026},
url={https://openreview.net/forum?id=IgzMfT6eUx}
}

@article{li2025lavida,
  title         = {{LaViDa: A Large Diffusion Language Model for Multimodal Understanding}},
  author        = {Li, Shufan and Kallidromitis, Konstantinos and Bansal, Hritik and Gokul, Akash and Kato, Yusuke and Kozuka, Kazuki and Kuen, Jason and Lin, Zhe and Chang, Kai-Wei and Grover, Aditya},
  journal       = {arXiv preprint arXiv:2505.16839},
  year          = {2025},
  eprint        = {2505.16839v4},
  archivePrefix = {arXiv},
  primaryClass  = {cs.CV},
  url           = {https://arxiv.org/abs/2505.16839v4}
}

@article{zhao2025d1,
  title         = {{d1: Scaling Reasoning in Diffusion Large Language Models via Reinforcement Learning}},
  author        = {Zhao, Siyan and Gupta, Devaansh and Zheng, Qinqing and Grover, Aditya},
  journal       = {arXiv preprint arXiv:2504.12216},
  year          = {2025},
  eprint        = {2504.12216v2},
  archivePrefix = {arXiv},
  primaryClass  = {cs.CL},
  url           = {https://arxiv.org/abs/2504.12216v2}
}

@article{hu2026residualcontextdiffusionlanguage,
  title         = {{Residual Context Diffusion Language Models}},
  author        = {Hu, Yuezhou and Singh, Harman and Maheswaran, Monishwaran and Xi, Haocheng and Hooper, Coleman and Zhang, Jintao and Tomar, Aditya and Mahoney, Michael W. and Min, Sewon and Farajtabar, Mehrdad and Keutzer, Kurt and Gholami, Amir and Xu, Chenfeng},
  journal       = {arXiv preprint arXiv:2601.22954},
  year          = {2026},
  eprint        = {2601.22954v2},
  archivePrefix = {arXiv},
  primaryClass  = {cs.CL},
  url           = {https://arxiv.org/abs/2601.22954v2}
}

@article{nara2026,
  title         = {{NaRA: Noise-Aware LoRA for Parameter-Efficient Fine-Tuning of Diffusion LLMs}},
  author        = {Wang, Shuaidi and Zhuang, Zhan and Huang, Ruping and Zhang, Yu},
  journal       = {arXiv preprint arXiv:2605.29716},
  year          = {2026},
  eprint        = {2605.29716v1},
  archivePrefix = {arXiv},
  primaryClass  = {cs.AI},
  url           = {https://arxiv.org/abs/2605.29716v1}
}

@article{cobbe2021gsm8k,
  title={Training Verifiers to Solve Math Word Problems},
  author={Cobbe, Karl and Kosaraju, Vineet and Bavarian, Mohammad and
          Chen, Mark and Jun, Heewoo and Kaiser, Lukasz and
          Plappert, Matthias and Tworek, Jerry and Hilton, Jacob and
          Nakano, Reiichiro and Hesse, Christopher and Schulman, John},
  journal={arXiv preprint arXiv:2110.14168},
  year={2021}
}

@inproceedings{hendrycks2021math,
  title={Measuring Mathematical Problem Solving With the {MATH} Dataset},
  author={Hendrycks, Dan and Burns, Collin and Kadavath, Saurav and
          Arora, Akul and Basart, Steven and Tang, Eric and
          Song, Dawn and Steinhardt, Jacob},
  booktitle={Proceedings of the Neural Information Processing Systems Track on Datasets and Benchmarks},
  volume={1},
  year={2021}
}

@inproceedings{huang2025opencoder,
  title={{O}pen{C}oder: The Open Cookbook for Top-Tier Code Large Language Models},
  author={Huang, Siming and Cheng, Tianhao and Liu, Jason Klein and
          Xu, Weidi and Hao, Jiaran and Song, Liuyihan and Xu, Yang and
          Yang, Jian and Liu, Jiaheng and Zhang, Chenchen and
          Chai, Linzheng and Yuan, Ruifeng and Luo, Xianzhen and
          Wang, Qiufeng and Fan, YuanTao and Zhu, Qingfu and
          Zhang, Zhaoxiang and Gao, Yang and Fu, Jie and Liu, Qian and
          Li, Houyi and Zhang, Ge and Qi, Yuan and Yinghui, Xu and
          Chu, Wei and Wang, Zili},
  booktitle={Proceedings of the 63rd Annual Meeting of the Association for Computational Linguistics (Volume 1: Long Papers)},
  pages={33167--33193},
  year={2025},
  doi={10.18653/v1/2025.acl-long.1591},
  url={https://aclanthology.org/2025.acl-long.1591/}
}

@article{wu2026regft,
  title={Learn Hard Problems During RL with Reference Guided Fine-tuning},
  author={Wu, Yangzhen and Li, Shanda and Wen, Zixin and Zhou, Xin and Talwalkar, Ameet and Yang, Yiming and Huang, Wenhao and Cai, Tianle},
  journal={arXiv preprint arXiv:2603.01223},
  year={2026}
}
\bibliographystyle{iclr2027_conference}

\appendix
\counterwithin{figure}{section}
\counterwithin{table}{section}

\newpage
\section{Additional Experiment Details}
\subsection{Training configurations}
Table~\ref{tab:training_hyperparameters} reports the SFT
hyperparameters. The shared GSM8K, MATH, and OpenCoder training corpus
is described in Section~\ref{sec:models_training_data}; PACS's
four-candidate canvas family with $B=32$ is defined in
Section~\ref{sec:experimental_setup}. Decoding settings and prompt
formats are given in Sections~\ref{app:prompt_decode}
and~\ref{app:generation_prompts}.

\begin{table}[H]
    \centering
    \caption{Hyperparameters used for PACS.}
    \label{tab:training_hyperparameters}
    \begin{tabular}{ll}
        \toprule
        \textbf{Hyperparameter} & \textbf{Value} \\
        \midrule
        Learning rate & $5\times10^{-6}$ \\
        Learning rate scheduler type & Cosine \\
        Adam $\beta$ parameters & $\beta_1=0.9,\ \beta_2=0.999$ \\
        Gradient accumulation steps & 4 \\
        Per-device train batch size & 2 \\
        Epochs & 1 \\
        Precision & bf16 \\
        LoRA $r$ & 32 \\
        LoRA $\alpha$ & 64 \\
        Weight decay & 0.0 \\
        Maximum gradient norm & 1.0 \\
        Warmup ratio & 5\% \\
        Random seed & 42 \\
        \bottomrule
    \end{tabular}
\end{table}
\label{app:exp_setup}

\subsection{Decode Configuration}
\label{app:prompt_decode}

The following settings are shared by all four benchmarks.
\begin{center}
\small
\setlength{\tabcolsep}{9pt}
\renewcommand{\arraystretch}{1.08}
\begin{tabular}{@{}ll@{}}
\toprule
\textbf{Parameter} & \textbf{Value} \\
\midrule
\texttt{gen\_length} & $\{128,256,512\}$ \\
\texttt{block\_length} & $32$ \\
\texttt{temperature} & $0.0$ (greedy) \\
\texttt{cfg\_scale} & $0.0$ \\
\texttt{confidence\_threshold} & $0.9$ \\
\texttt{seed} & $42$ \\
\bottomrule
\end{tabular}
\end{center}
At temperature zero, \nolinkurl{add_gumbel_noise} is a no-op.
Decoding uses {blockwise confidence-threshold parallel unmasking}:
at each forward pass, every masked position in the active block whose
top-1 softmax probability is at least $0.9$ is committed. If no position
meets the threshold, the single highest-confidence masked position is
committed \citep{wu2025fastdllm}.

The same confidence-guided selection rule is used to construct PACS's
reference-conditioned training trajectories: the model determines
which positions are revealed at each step, while the reference supplies
their token values.
\clearpage

\subsection{Generation Prompts}
\label{app:generation_prompts}

All four benchmarks use a single user turn, with no added task instruction
or few-shot exemplars.
\begingroup
\emergencystretch=2em
\lstset{
    language={},
    basicstyle=\fontencoding{T1}\ttfamily\fontsize{8}{9.5}\selectfont,
    columns=fullflexible,keepspaces=true,
    showstringspaces=false,showspaces=false,showtabs=false,
    upquote=true,tabsize=4,
    breaklines=true,breakatwhitespace=false,breakindent=0pt,
    postbreak=\mbox{\textcolor{gray}{\tiny$\hookrightarrow$}\space},
    frame=single,rulecolor=\color{black!20},
    backgroundcolor=\color{black!2},framesep=4pt,
    xleftmargin=4pt,xrightmargin=4pt,
    aboveskip=5pt,belowskip=7pt,
    numbers=none
}
\begin{lstlisting}
msgs = [{"role": "user", "content": benchmark_item}]
s = tokenizer.apply_chat_template(msgs, add_generation_prompt=True, tokenize=False)
\end{lstlisting}

\subsubsection{HumanEval}
\label{app:prompt_humaneval}
\paragraph{LLaDA-8B-Instruct}
Source:

\begin{lstlisting}
<|startoftext|><|start_header_id|>user<|end_header_id|>

from typing import List

def has_close_elements(numbers: List[float], threshold: float) -> bool:
    """ Check if in given list of numbers, are any two numbers closer to each other than
    given threshold.
    >>> has_close_elements([1.0, 2.0, 3.0], 0.5)
    False
    >>> has_close_elements([1.0, 2.8, 3.0, 4.0, 5.0, 2.0], 0.3)
    True
    """<|eot_id|><|start_header_id|>assistant<|end_header_id|>

\end{lstlisting}
\subsubsection{MBPP}
\label{app:prompt_mbpp}

\medskip\noindent\textbf{LLaDA-8B-Instruct}\par\nobreak
\begin{lstlisting}
<|startoftext|><|start_header_id|>user<|end_header_id|>

def remove_Occ(s,ch):
    """Write a python function to remove first and last occurrence of a given character from the string.
    
    >>> remove_Occ('hello', 'l')
    'heo'
    >>> remove_Occ('abcda', 'a')
    'bcd'
    >>> remove_Occ('PHP', 'P')
    'H'
    """<|eot_id|><|start_header_id|>assistant<|end_header_id|>

\end{lstlisting}
\endgroup

\clearpage
\subsection{Selected-Bin Distributions}
\label{app:selected_bin_distributions}

\begin{figure}[!ht]
    \centering
    \includegraphics[width=\textwidth]{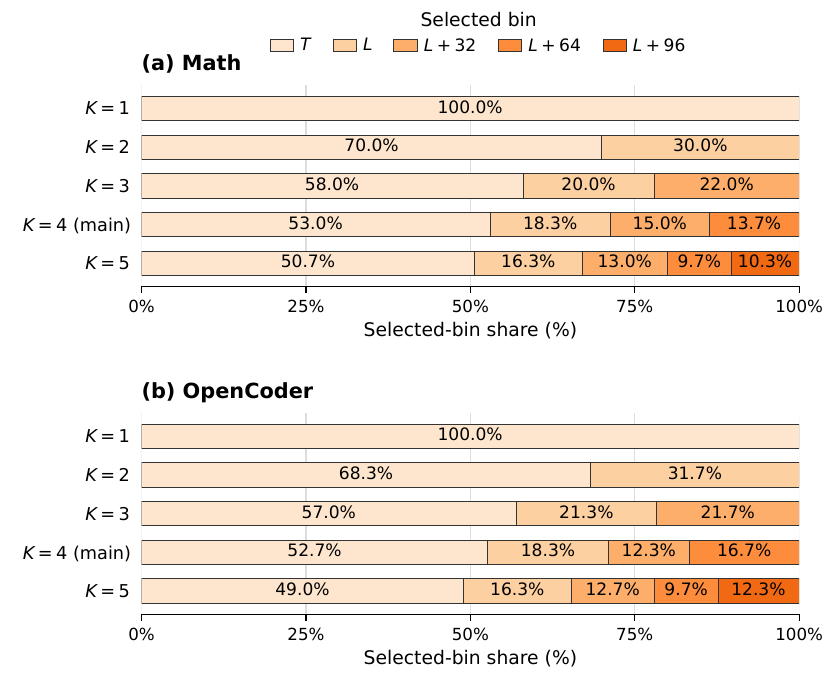}
    \caption{\textbf{Selected-bin proportions across candidate families.}
Each horizontal bar is normalized independently to $100\%$ and shows
the fraction of examples for which PACS selects each candidate in the
nested family $\Lambda_K(P,A)$. We use
$L_{\mathrm{ref}}=|P|+|A|$,
$L_{\min}=B\left\lceil L_{\mathrm{ref}}/B\right\rceil$, and $B=32$,
so the bins correspond to
$L_{\mathrm{ref}}$, $L_{\min}$, $L_{\min}+B$,
$L_{\min}+2B$, and $L_{\min}+3B$.
Segment labels give percentages rounded to one decimal place.
Bins not contained in a given $\Lambda_K$ have no segment, and colors
identify the same candidate lengths across datasets and families.
$\Lambda_4$ is the main configuration.}
    \label{fig:selected_bin_proportions}
\end{figure}

Figure~\ref{fig:selected_bin_proportions} shows which canvas bins are
selected within each cumulative candidate family for Math and OpenCoder.
Let $T=|P|+|A|$ and $L=L_{\min}(P,A)$. The families are
\[
\begin{aligned}
\Lambda_1 &= \{T\},\\
\Lambda_K &= \{T\}\cup\{L+32j\mid j=0,\ldots,K-2\},
\qquad K\in\{2,3,4,5\}.
\end{aligned}
\]
Each family retains all earlier candidates; duplicate lengths are
considered once. Here $K$ indexes the nominal family, not its cardinality
after deduplication, and $K=4$ is the main configuration.

The selected bins show that predictability-based canvas selection does
not collapse to the shortest feasible canvas or to a single fixed
offset. Under our main family $\Lambda_4$, PACS selects
$L_{\mathrm{ref}}$ for only $53.0\%$ of Math and $52.7\%$ of
OpenCoder examples; the remaining $47.0\%$ and $47.3\%$, respectively,
are distributed across $L_{\min}$, $L_{\min}+B$, and
$L_{\min}+2B$. Thus, for nearly half of the training examples, the
reference is better supported on a canvas extending beyond its exact
length, and the preferred extent varies across examples. This
heterogeneity is precisely why a single global canvas rule is
insufficient for our setting: exploiting the candidate family requires
selecting, for each reference, the canvas on which its commitment-level
predictions are best supported. PACS performs this selection using
reference predictability rather than imposing the same extent on every
example. The similar distributions across Math and OpenCoder suggest
that this need for example-specific canvas selection is not confined to
one domain. Finally, expanding to $\Lambda_5$ assigns only $10.3\%$ of
Math and $12.3\%$ of OpenCoder examples to the newly added
$L_{\min}+3B$ candidate, consistent with diminishing returns from
searching substantially beyond the main family.

\begingroup
\raggedbottom
\subsection{Complete Results Across Generation Lengths}
\label{app:full_benchmark_results}
Tables~\ref{tab:full_llada_results} and~\ref{tab:full_dream_results}
report the complete $K_{\mathrm{eval}}\in\{128,256,512\}$ comparisons.
$\Delta$ is PACS's relative percentage gain over
the corresponding initial checkpoint.

\begin{table}[H]
    \centering
    \caption{Complete LLaDA-8B-Instruct results (\%) across $K_{\mathrm{eval}}\in\{128,256,512\}$.}
    \label{tab:full_llada_results}
    \begingroup
    \normalsize
    \setlength{\tabcolsep}{3pt}
    \renewcommand{\arraystretch}{1.05}
    \begin{tabular*}{\textwidth}{@{\extracolsep{\fill}}lrrr@{\hspace{12pt}}rrr@{}}
        \toprule
        & \multicolumn{3}{c}{\textbf{GSM8K}} & \multicolumn{3}{c}{\textbf{MATH500}} \\
        \cmidrule(lr){2-4}\cmidrule(l){5-7}
        \textbf{Model / Seq Len} & \textbf{128} & \textbf{256} & \textbf{512} & \textbf{128} & \textbf{256} & \textbf{512} \\
        \midrule
        \textbf{LLaDA-Instruct} & 76.12 & 75.44 & 79.08 & 35.60 & 36.40 & 38.80 \\
        \midrule
        + Vanilla & 77.03 & 76.27 & 78.77 & 35.60 & 37.00 & 39.40 \\
        + Semi-AR Block & 76.12 & \textbf{78.20} & 79.23& 34.00 & 38.40 & 39.20 \\
        + LIFT & 75.89 & 77.63 & 78.85 & 35.40 & 37.00 & 38.00 \\
        + TRIMS & 76.88 & 77.63 & 79.53 & 34.80 & 38.00 & 39.60 \\
        \midrule
        + PACS (Ours) & \textbf{77.10} & 77.94 & \textbf{80.97} & \textbf{35.80} & \textbf{39.00} & \textbf{40.00} \\
        \midrule
        $\Delta (\%)$ & \textcolor{green!45!black}{+1.29\%} & \textcolor{green!45!black}{+3.31\%} & \textcolor{green!45!black}{+2.39\%} & \textcolor{green!45!black}{+0.56\%} & \textcolor{green!45!black}{+7.14\%} & \textcolor{green!45!black}{+3.09\%} \\
        \bottomrule
    \end{tabular*}
    \par\medskip
    \begin{tabular*}{\textwidth}{@{\extracolsep{\fill}}lrrr@{\hspace{12pt}}rrr@{}}
        \toprule
        & \multicolumn{3}{c}{\textbf{MBPP}} & \multicolumn{3}{c}{\textbf{HumanEval}} \\
        \cmidrule(lr){2-4}\cmidrule(l){5-7}
        \textbf{Model / Seq Len} & \textbf{128} & \textbf{256} & \textbf{512} & \textbf{128} & \textbf{256} & \textbf{512} \\
        \midrule
        \textbf{LLaDA-Instruct} & 24.12 & 40.86 & 49.81 & 28.05 & 37.80 & 43.90 \\
        \midrule
        + Vanilla & 24.90 & 49.03 & 50.19 & 26.83 & 38.41 & 46.95 \\
        + Semi-AR Block & 34.24 & 48.60 & 50.20 & 27.40 & 43.90 & \textbf{48.20} \\
        + LIFT & 27.63 & 46.69 & 48.25 & 25.00 & 39.63 & 46.30 \\
        + TRIMS & 34.24 & \textbf{52.92} & 50.58 & 23.78 & 40.24 & 45.73 \\
        \midrule
        + PACS (Ours) & \textbf{37.74} & 49.42 & \textbf{52.14} & \textbf{30.49} & \textbf{47.56} & 46.95 \\
        \midrule
        $\Delta (\%)$ & \textcolor{green!45!black}{+56.47\%} & \textcolor{green!45!black}{+20.95\%} & \textcolor{green!45!black}{+4.68\%} & \textcolor{green!45!black}{+8.70\%} & \textcolor{green!45!black}{+25.82\%} & \textcolor{green!45!black}{+6.95\%} \\
        \bottomrule
    \end{tabular*}
    \endgroup
\end{table}

\begin{table}[H]
    \centering
    \caption{Complete Dream-7B-Instruct results (\%) across $K_{\mathrm{eval}}\in\{128,256,512\}$.}
    \label{tab:full_dream_results}
    \begingroup
    \normalsize
    \setlength{\tabcolsep}{3pt}
    \renewcommand{\arraystretch}{1.05}
    \begin{tabular*}{\textwidth}{@{\extracolsep{\fill}}lrrr@{\hspace{12pt}}rrr@{}}
        \toprule
        & \multicolumn{3}{c}{\textbf{GSM8K}} & \multicolumn{3}{c}{\textbf{MATH500}} \\
        \cmidrule(lr){2-4}\cmidrule(l){5-7}
        \textbf{Model / Seq Len} & \textbf{128} & \textbf{256} & \textbf{512} & \textbf{128} & \textbf{256} & \textbf{512} \\
        \midrule
        \textbf{Dream-Instruct-7B} & 70.66 & \textbf{80.52} & 80.74 & {37.20} & 43.80 & 44.20 \\
        \midrule
        + Vanilla & 70.35 & 80.06 & 80.36 & \textbf{37.20} & 44.20 & 44.80 \\
        + Semi-AR Block & 70.43 & 80.21 & 80.06 & 37.00 & 43.60 & 43.40 \\
        + LIFT & 71.41 & 80.36 & 80.81 & 36.00 & 44.80 & 45.20 \\
        + TRIMS & 71.26 & 80.44 & 80.59 & 37.40 & 45.60 & 45.40 \\
        \midrule
        + PACS (Ours) & \textbf{72.33} & \textbf{80.52} & \textbf{80.89} & \textbf{37.20} & \textbf{49.20} & \textbf{46.20} \\
        \midrule
        $\Delta$ (\%) & \textcolor{green!45!black}{+2.36\%} & \textcolor{green!45!black}{0.00\%} & \textcolor{green!45!black}{+0.19\%} & \textcolor{green!45!black}{0.00\%} & \textcolor{green!45!black}{+12.33\%} & \textcolor{green!45!black}{+4.52\%} \\
        \bottomrule
    \end{tabular*}
    \par\medskip
    \begin{tabular*}{\textwidth}{@{\extracolsep{\fill}}lrrr@{\hspace{12pt}}rrr@{}}
        \toprule
        & \multicolumn{3}{c}{\textbf{MBPP}} & \multicolumn{3}{c}{\textbf{HumanEval}} \\
        \cmidrule(lr){2-4}\cmidrule(l){5-7}
        \textbf{Model / Seq Len} & \textbf{128} & \textbf{256} & \textbf{512} & \textbf{128} & \textbf{256} & \textbf{512} \\
        \midrule
        \textbf{Dream-Instruct-7B} & 31.52 & 45.53 & 46.69 & 45.73 & 51.83 & 48.17 \\
        \midrule
        + Vanilla & 48.25 & 60.31 & 59.92 & 42.07 & 54.27 & 50.61 \\
        + Semi-AR Block & \textbf{49.81} & 61.09 & 58.75 & 45.73 & 58.53 & 56.71 \\
        + LIFT & 33.07 & 50.58 & 51.75 & 46.34 & 52.44 & 54.26 \\
        + TRIMS & 32.30 & 52.14 & 51.36 & 41.46 & 53.66 & 56.10 \\
        \midrule
        + PACS (Ours) & 48.64 & \textbf{61.48} & \textbf{61.48} & \textbf{47.56} & \textbf{60.98} & \textbf{57.93} \\
        \midrule
        $\Delta$ (\%) & \textcolor{green!45!black}{+54.31\%} & \textcolor{green!45!black}{+35.03\%} & \textcolor{green!45!black}{+31.68\%} & \textcolor{green!45!black}{+4.00\%} & \textcolor{green!45!black}{+17.65\%} & \textcolor{green!45!black}{+20.26\%} \\
        \bottomrule
    \end{tabular*}
    \endgroup
\end{table}
\clearpage
\endgroup

\clearpage
\section{Supplementary Experiments}
\label{app:supplementary}

\subsection{Robustness to training datasets}
\label{app:alt_corpus}
To test whether the gains of PACS depend on the training-data source,
we repeat the comparison using an alternative mathematics--code corpus.
For mathematical reasoning, we use Math14k, following the training setup
of NaRA \citep{nara2026}; Math14k combines GSM8K and AQuA examples with
model-generated rationales.
For code generation, we use KodCode, a large-scale synthetic coding
dataset with verified solutions and unit tests \citep{xu2025kodcode}.

All methods are trained on the same retained examples and data splits
under this alternative corpus, and are evaluated with the same decoding
configuration as in the main experiments.
This comparison tests whether the relative benefit of PACS persists
when the supervision data are changed, rather than being specific to
the training sources used in the main results.

\begin{table}[H]
\centering
\caption{Performance on the alternative training corpus.
All methods use Math14k and KodCode and are evaluated using
LLaDA-8B-Instruct at $K_{\mathrm{eval}}=256$.
Bold marks the highest score in each benchmark column.}
\label{tab:alt_corpus}
\begin{tabular*}{\linewidth}{@{\extracolsep{\fill}}lrrrr@{}}
\toprule
Method & GSM8K & MATH500 & MBPP & HumanEval \\
\midrule
Vanilla SFT   & 75.82 & 36.20 & 42.41 & 36.59 \\
Semi-AR block & \textbf{77.56} & 37.60 & 41.63 & 37.80 \\
LIFT          & 77.10 & 36.40 & 40.08 & 38.41 \\
TRIMS         & 77.03 & 37.20 & \textbf{43.58} & 35.98 \\
PACS          & 77.41 & \textbf{38.20} & 42.80 & \textbf{40.24} \\
\bottomrule
\end{tabular*}
\end{table}

PACS improves over vanilla SFT on all four benchmarks and achieves the highest accuracy on MATH500 and HumanEval under the alternative training corpus, supporting the persistence of its benefits across training-data sources.

\subsection{Sanity Reproduction of Prior Baselines}
\label{app:sanity_reproduction}

To validate our implementations of the strongest SFT baselines, we
reproduce representative results from LIFT and TRIMS using
LLaDA-8B-Instruct.
For each method, we follow the corresponding paper's training data,
number of fine-tuning epochs, and evaluation configuration rather than
our PACS training recipe.
For LIFT, we reproduce the reported $\mathrm{LIFT}_3$ results on
GSM8K and MATH500 using the 20-epoch training setup and
$K_{\mathrm{eval}}=512$ \citep{parashar2026lift}.
For TRIMS, we reproduce its coding results on MBPP and HumanEval using
the reported 30-epoch training recipe and the same decoding
configuration as the original evaluation \citep{chen2026trims}.

\begin{table}[H]
\centering
\caption{
Sanity reproduction of prior SFT baselines on LLaDA-8B-Instruct.
Reported values are taken from the corresponding papers; each
reproduction follows the training and evaluation configuration of the
original result.
}
\label{tab:sanity_reproduction}
\small
\setlength{\tabcolsep}{5pt}
\renewcommand{\arraystretch}{1.08}
\begin{tabular}{llrr}
\toprule
\textbf{Method} & \textbf{Benchmark}
& \textbf{Reported} & \textbf{Reproduction} \\
\midrule
LIFT$_3$ & GSM8K     & 79.40 & 79.30 \\
LIFT$_3$ & MATH500   & 38.40 & 38.20 \\
\midrule
TRIMS     & MBPP      & 41.80 & 41.63 \\
TRIMS     & HumanEval & 37.20 & 36.59 \\
\bottomrule
\end{tabular}
\end{table}

The reproduced scores are within 0.61 percentage points of the
corresponding reported results across all four comparisons
(Table~\ref{tab:sanity_reproduction}). This close agreement provides
a sanity check for our implementations before the controlled
comparisons in the main experiments.

For the controlled comparisons in the main experiments, we retain each
baseline's paper-selected training configuration rather than retuning it
under the PACS setup.
For LIFT, we use the $\mathrm{LIFT}_3$ variant reported as the strongest
configuration in the original study; TRIMS likewise uses the
configuration selected in its paper.
This keeps the baseline comparisons tied to the methods' reported
training choices while changing only the shared data and evaluation
protocol used in our study.

\subsection{Computational Cost}
\label{app:compute_cost}
\label{sec:cost_robustness}

All methods use a common prepared corpus, whose construction cost is
excluded from the runtime comparison.
Table~\ref{tab:runtime} summarizes runtime on the OpenCoder workload.
Vanilla and semi-autoregressive SFT use 20-epoch schedules.
For TRIMS and LIFT, we adopt the 30-epoch and 20-epoch main training
schedules reported in their respective papers
\citep{chen2026trims,parashar2026lift}.
Vanilla SFT samples the masking probability
$\alpha\sim\operatorname{Unif}(0.1,1)$.

\begin{table}[H]
    \centering
    \caption{
    Runtime on the OpenCoder workload.
    }
    \label{tab:runtime}
    \small
    \setlength{\tabcolsep}{12pt}
    \renewcommand{\arraystretch}{1.10}
    \begin{tabular}{@{}ll@{}}
        \toprule
        \textbf{Method} & \textbf{Runtime} \\
        \midrule
        Vanilla SFT   & $\approx$1\,h\,13\,min \\
        Semi-AR block & $\approx$1\,h\,16\,min \\
        LIFT          & $\approx$8\,h\,24\,min \\
        TRIMS         & $\approx$1\,h\,21\,min \\
        PACS          & $\approx$14\,h\,55\,min \\
        \bottomrule
    \end{tabular}

    \vspace{4pt}
    \begin{minipage}{0.95\linewidth}
        \footnotesize
        Timings use one H200.
        The timed LIFT variant is LIFT$_3$. The timed PACS run
        uses four candidate lengths.
    \end{minipage}
\end{table}

PACS allocates additional computation to constructing and learning
from successive reference-conditioned prediction states.
Confidence-guided position selection is recomputed as reference
tokens are revealed, and canvas selection compares this process
across candidate lengths.
Its higher computational cost therefore reflects multiple prediction
states per reference, rather than simply more epochs of independently
sampled masking configurations.
\paragraph{Matched post-training budgets.}
To distinguish the effect of the supervision procedure from that of additional computation, the matched-budget comparison includes both method-specific preprocessing and optimization, while excluding shared
corpus construction.
Comparisons use common hardware and validation-selected checkpoints,
evaluated at $K_{\mathrm{eval}}=256$.
Table~\ref{tab:compute_matched} reports the results.

\begin{table}[H]
\centering
\caption{
Compute-matched comparison on
LLaDA-8B-Instruct at $K_{\mathrm{eval}}=256$.
Budgets include method-specific preprocessing and optimization,
excluding shared corpus construction.
Accuracy is reported on MATH500 ($n=500$) and HumanEval ($n=164$).
Bold marks the highest score for each benchmark and budget.
}
\label{tab:compute_matched}
\begingroup
\small
\setlength{\tabcolsep}{3pt}
\renewcommand{\arraystretch}{1.15}
\begin{tabular*}{\linewidth}
    {@{\extracolsep{\fill}}lrrrrrr@{}}
\toprule
& \multicolumn{2}{c}{4 GPU-hours}
& \multicolumn{2}{c}{8 GPU-hours}
& \multicolumn{2}{c}{12 GPU-hours} \\
\cmidrule(lr){2-3}
\cmidrule(lr){4-5}
\cmidrule(lr){6-7}
Method
& MATH500 & HumanEval
& MATH500 & HumanEval
& MATH500 & HumanEval \\
\midrule
Vanilla SFT
& 37.20 & 39.02 & 37.40 & 39.63 & 37.20 & 39.02 \\
Semi-AR block
& \textbf{38.40} & \textbf{43.90} & 38.60 & \textbf{44.51} & 38.60 & 43.90 \\
LIFT
& 36.40 & 38.41 & 37.00 & 39.63 & 37.20 & 40.85 \\
TRIMS
& 37.60 & 40.24 & 38.20 & 40.85 & 38.20 & 39.63 \\
PACS
& 37.40 & 40.24 & \textbf{38.80} & 43.90 & \textbf{39.00} & \textbf{46.34} \\
\bottomrule
\end{tabular*}
\endgroup
\end{table}
Table~\ref{tab:compute_matched} shows that PACS makes effective use of
additional post-training computation. It achieves the highest MATH500
accuracy at 8 GPU-hours and leads both benchmarks at 12 GPU-hours,
surpassing the strongest baseline by 0.40 percentage points on MATH500
and 2.44 points on HumanEval. These gains under matched total budgets
support the effectiveness of PACS beyond simply allocating more
computation. Its growing advantage at larger budgets highlights the
value of directing post-training computation toward predictability-aware
canvas selection and supervision.

\paragraph{Effect of the training-time confidence threshold.}
The training confidence threshold offers a way to trade off training cost
against downstream accuracy. A lower threshold allows PACS to reveal more
reference tokens per forward pass, which can reduce the number of trajectory
steps and pre-commitment states used for supervision.
Table~\ref{tab:pacs-compute-threshold} compares partial and full PACS runs at
a training threshold of $0.2$ with semi-AR block SFT and the main PACS
configuration at $0.9$.

\begin{table}[t]
    \centering
    \small
    \caption{Training time and coding accuracy (\%) for LLaDA-8B-Instruct
    at $K_{\mathrm{eval}}=256$. Times are for one H200.}
    \label{tab:pacs-compute-threshold}
    \begin{tabular}{lcrrr}
        \toprule
        Configuration & $\tau_{\mathrm{train}}$ & Time & HumanEval & MBPP \\
        \midrule
        Semi-AR block & -- & $\approx$1 h 16 min & 43.90 & 48.60 \\
        PACS (full run) & 0.9 & $\approx$14 h 55 min & 47.56 & 49.42 \\
        \midrule
        PACS (partial run) & 0.2 & $\approx$1 h 5 min & 44.51 & 48.60 \\
        PACS (full run) & 0.2 & $\approx$3 h 15 min & 45.73 & 49.01 \\
        \bottomrule
    \end{tabular}
\end{table}

At a comparable training time, the partial PACS run
scores higher than the full semi-AR run on HumanEval
and matches it on MBPP. Completing the lower-threshold run improves both
scores over the partial run and takes about 22\% of the training time of the
full $0.9$ run. The higher-threshold run still achieves the best accuracy on
both benchmarks. These results illustrate how the threshold can be adjusted
to fit different training budgets.

\subsection{Matched-State Predictive Entropy}
\label{app:matched-state-entropy}

To test whether PACS produces more concentrated token predictions than other
SFT methods, we evaluate the LLaDA-8B-Instruct checkpoints on the same 614
reference-conditioned block-entry states from HumanEval and MBPP. We use a 256-token generation canvas and a block size of
32. In each state, earlier reference blocks are visible, while the current and
future blocks remain masked. We compute full-vocabulary token entropy at
temperature~1 for reference positions in the current block, excluding padding.
We average over scored tokens within each problem, average problems within
each benchmark, and then give the two benchmark means equal weight.

\begin{table}[t]
    \centering
    \small
    \caption{Mean predictive token entropy on shared reference-conditioned
    block-entry states.}
    \label{tab:matched-state-entropy}
    \begin{tabular}{lr}
        \toprule
        Method & Matched-state token entropy (nats) \\
        \midrule
        Naive SFT   & 3.2246 \\
        Semi-AR SFT & 2.9480 \\
        PACS        & 3.3177 \\
        \bottomrule
    \end{tabular}
\end{table}

PACS has higher predictive entropy than both SFT controls on these shared
states (Table~\ref{tab:matched-state-entropy}), which suggests that its accuracy gains do not come from sharpening its predictive distribution, a pattern often associated with reduced exploration in subsequent RL.

\subsection{Canvas Selection under Other Supervision Rules}
\label{app:other_supervision}
\begin{figure}[H]
    \centering
    \includegraphics[width=\textwidth]{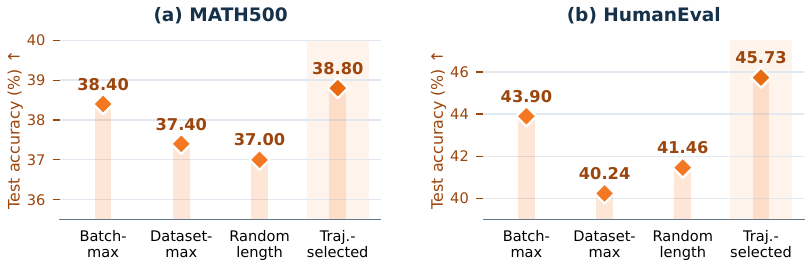}
    \caption{\textbf{Semi-AR block canvas-choice test accuracy.}
    Post-SFT test accuracy (orange diamonds) on MATH500 and HumanEval at
    $K_{\mathrm{eval}}=256$ for four canvas-length rules.}
    \label{fig:semiar_canvas_accuracy}
\end{figure}

Within each supervision pair, we hold the references, loss support, normalization, and optimization fixed while changing only the canvas rule.
Table~\ref{tab:other_supervision} compares batch-max and PACS-selected lengths
under random-mask SFT, semi-AR block SFT, and confidence-guided reference
revelation. Figure~\ref{fig:semiar_canvas_accuracy} further compares
four canvas-length rules under semi-AR block SFT.

\begin{table}[H]
\centering
\caption{
Canvas selection under different supervision rules.
LLaDA-8B-Instruct at $K_{\mathrm{eval}}=256$. For trajectory-selected rows, canvas lengths are chosen from the same
four-candidate family used in the main experiments.
}
\label{tab:other_supervision}

\begingroup
\small
\setlength{\tabcolsep}{2pt}
\renewcommand{\arraystretch}{1.15}

\begin{tabular*}{\linewidth}
{@{\extracolsep{\fill}}llrrrr@{}}
\toprule
Supervision & Length rule
& GSM8K & MATH500 & MBPP & HumanEval \\
\midrule

Random-mask SFT
& Batch-max
& 76.27 & 37.00 & {49.03} & 38.41 \\

Random-mask SFT
& Trajectory-selected
& \textbf{76.50} & \textbf{37.20} & \textbf{49.42} & \textbf{39.02} \\

\midrule

Semi-AR block
& Batch-max
& 78.20 & 38.40 & 48.60 & 43.90 \\

Semi-AR block
& Trajectory-selected
& \textbf{78.39} & \textbf{38.80} & \textbf{49.03} & \textbf{45.73} \\

\midrule

Confidence trajectory
& Batch-max
& 77.48 & 38.00 & 47.08 & 40.90 \\

Confidence trajectory
& Trajectory-selected
& \textbf{77.94} & \textbf{39.00} & \textbf{49.42} & \textbf{47.56} \\

\bottomrule
\end{tabular*}

\endgroup
\end{table}

PACS-selected canvases improve or match batch-max accuracy on all
four benchmarks under each supervision rule.
On HumanEval, the gains increase from 0.61 percentage points under
random masking to 1.83 under semi-AR block SFT and 6.66 under
confidence-guided reference revelation.
The four-rule semi-AR comparison further shows that selected canvases
outperform batch-max, dataset-max, and random lengths on both
MATH500 and HumanEval.
These results support the applicability of predictability-based
canvas selection across supervision rules, with the largest observed
gains when paired with confidence-guided reference revelation.

\section{Details of the Canvas Diagnostics}
\label{app:sec3exp}

\paragraph{Experimental settings.}
We use LLaDA-8B-Instruct on 300 prompts from GSM8K.
Reference-conditioned trajectories follow the confidence-guided
decoding rule in Section~\ref{sec:confidence_decoding}, with confidence threshold $\tau=0.9$
and block size $B=32$.
Source responses are generated at temperature $0$.

\subsection{Interventions on Revealed Content}
\label{app:content_interventions}

Both interventions start from a target's original pre-commitment
canvas.
We leave the target masked and modify only the specified surrounding
reference content.
The target is then reevaluated on the modified canvas; its identity
and reference value remain unchanged. For each candidate generation length $K$, we retain only examples
whose scored reference length satisfies $m<K$.
Both the candidate and source-length evaluations use this same
subset and score the same complete reference spans.
The retained subset may therefore vary across candidate lengths.

\paragraph{Withholding established content.}
We consider targets whose original commitment predictions agree with
the reference.
For a target selected at iteration $s$, we replace the reference
tokens revealed at iteration $s-1$ with $\mathtt{[MASK]}$,
leaving all other positions unchanged.
We restrict the comparison to targets that were already eligible
at iteration $s-1$.
This restriction excludes targets that become eligible only after
the preceding step, for example when decoding advances to a new block.

\paragraph{Revealing subsequent content.}
We consider targets whose original commitment predictions disagree
with the reference.
Starting from the original pre-commitment canvas, we reveal exactly
one reference token randomly selected at iteration $s+1$ of the original trajectory, keeping the current target masked.
All other positions remain unchanged.
The next step is defined by the original reveal schedule, not by
sequence position.

\paragraph{Reported quantities.}
For each intervention, we report the fraction of targets whose
reference-agreement status changes and the mean paired CE difference.
The difference is computed as the intervention CE minus the original
commitment CE.
The two interventions use different initial cohorts:
reference-agreeing targets for withholding and
reference-disagreeing targets for additional revelation.
Confidence intervals are obtained by prompt-level bootstrap
resampling.

\subsection{Interventions on Canvas Extent}
\label{app:extent_interventions}

The frozen model first generates each reference response at source
generation length $K_0=384$.
We then keep the reference and model parameters fixed while varying
the evaluation extent $K$, corresponding to total canvas length
$L=|P|+K$.

\paragraph{Fixed-trajectory control.}
We retain the source reveal trajectory and reevaluate its
pre-commitment predictions after changing the attention-visible
canvas extent.
The reveal schedule is not recomputed.
This control examines the effect of extent without allowing
confidence-based selection to produce a different trajectory.

\paragraph{Candidate-trajectory comparison.}
For each candidate length, we construct a new reference-conditioned trajectory.
Both the predictions and the subsequent selections can therefore
change with the canvas extent.
We score each reference target on its pre-commitment canvas and
aggregate using Equation~\ref{eq:trace_tight_score}.

For each construction, Figure~\ref{fig:blockce}A reports the change
in mean CE relative to the source-length evaluation.
Figure~\ref{fig:blockce}B reports the corresponding position-wise
changes for candidate trajectories at $K=256$ and $K=512$.
The increase toward later response positions is consistent with the
continuation and termination effects illustrated in
Figure~\ref{fig:intro}B.

\subsection{Schedule-Matched Random-Order Control}
\label{app:random_order}

We construct a random-order trajectory that matches the reveal schedule of
the corresponding PACS trajectory while removing confidence-based
position selection.

For block $b$ and step $s$, let
\[
q_{b,s}
=
\left|S^{\mathrm{PACS}}_{b,s}\right|
\]
denote the number of positions selected by PACS at that step. Let
$\mathcal E^{\mathrm{rand}}_{b,s}$ be the currently eligible masked
positions in the active block of the random trajectory. We sample
\begin{equation}
S^{\mathrm{rand}}_{b,s}
\sim
\operatorname{Unif}
\left\{
S\subseteq\mathcal E^{\mathrm{rand}}_{b,s}
:
|S|=q_{b,s}
\right\}.
\label{eq:random_schedule_match}
\end{equation}
Thus, PACS and the random-order control reveal the same number of positions
at every step, but differ in which eligible positions are selected.

For each $i\in S^{\mathrm{rand}}_{b,s}$, the reference-token loss is
computed on the pre-revelation canvas. The selected masks are then replaced
by their corresponding reference values before the next step:
\begin{equation}
C^{(s+1)}_i
=
\begin{cases}
y_i, & i\in S^{\mathrm{rand}}_{b,s},\\
C^{(s)}_i, & \text{otherwise}.
\end{cases}
\end{equation}
Consequently, both variants use reference-consistent intermediate canvases
and identical per-step reveal counts; the only difference is whether the
revealed positions are chosen by model confidence or uniformly at random.

Equivalently, the random trajectory can be constructed by sampling one
uniform random permutation of the eligible positions within each block and
partitioning that permutation according to the PACS reveal counts
$(q_{b,1},q_{b,2},\ldots)$. This implements sampling without replacement
while exactly preserving the number of reveals at each step.

This control is intentionally schedule matched rather than an independent
random decoding rule: its purpose is to isolate the contribution of
confidence-based position ordering from differences in reveal rate,
trajectory length, or amount of visible reference context.

\section{Algorithm}
\begin{algorithm}[H]
\caption{Canvas selection and commitment-level SFT}
\begin{algorithmic}[1]
\Require Training pairs $\mathcal D$, initial model $p_{\theta_0}$,
         candidate sets $\Lambda(P,A)$
\State Initialize $p_\theta$ from $p_{\theta_0}$
\For{each training minibatch $\mathcal B\subseteq\mathcal D$}
    \ForAll{$(P,A)\in\mathcal B$}
        \State Determine the scored reference span $R$ and $m=|R|$
        \ForAll{$L\in\Lambda(P,A)$}
            \State Construct $\mathcal T_L(P,A)$ using the current
                   $p_\theta$ with confidence-guided reference revelation
                   and no gradient tracking
            \State Compute $\operatorname{CE}_{\mathrm{tr}}(P,A;L)$
        \EndFor
        \State Select $L^*(P,A)$ using \eqref{eq:selected_training_length}
    \EndFor
    \State Keep the selected canvas lengths fixed for $\mathcal B$
    \State Fine-tune $p_\theta$ on $\mathcal B$ using \eqref{eq:commitment_sft_objective}
\EndFor
\State \Return $p_\theta$
\end{algorithmic}
\end{algorithm}

\end{document}